\documentclass[journal]{IEEEtran}
\usepackage{amsmath,amsfonts}
\usepackage{algorithmic}
\usepackage{algorithm}
\usepackage{array}
\usepackage[caption=false,font=normalsize,labelfont=sf,textfont=sf]{subfig}
\usepackage{textcomp}
\usepackage{stfloats}
\usepackage{url}
\usepackage{verbatim}
\usepackage{graphicx}
\usepackage{cite}

\usepackage{amsmath}
\usepackage{booktabs} 
\usepackage{multirow}
\usepackage{pifont}
\usepackage{amssymb}
\usepackage{xcolor}
\usepackage[table]{xcolor}
\usepackage{placeins}
\newcommand{\gr}[1]{{\color{black}#1}}
\newcommand{\eg}{\textit{e}.\textit{g}.}

\newcommand{\etal}{\textit{et al. }}

\usepackage{hyperref}
\hypersetup{
    colorlinks=true,
    linkcolor=blue,
    citecolor=blue,
    urlcolor=magenta
}

\begin{document}

\title{Bridging Adversarial and Collaborative Learning for AI-Generated Image Quality Assessment

\author{Baoliang Chen*, Qing Lin*, and Sijie Mai}
\IEEEcompsocitemizethanks{
\IEEEcompsocthanksitem *These authors contributed equally.
\IEEEcompsocthanksitem Baoliang Chen and Sijie Mai are with the Department of Computer Science, South China Normal University, China. Emails: blchen@scnu.edu.cn; sijiemai@m.scnu.edu.cn.
\IEEEcompsocthanksitem Qing Lin is with the Department of Data Science and Analytics, Hong Kong University of Science and Technology (Guangzhou). Email: qlin088@connect.hkust-gz.edu.cn.
\IEEEcompsocthanksitem Corresponding author: Sijie Mai
\IEEEcompsocthanksitem This work was supported by the National Natural Science Foundation of China under Grant No. 62401214 and the Guangdong Philosophy and Social Sciences Planning Project under Grant No. GD26YJY34.
}
}

\maketitle


\maketitle
\begin{abstract}
AI-generated image quality assessment (AIGIQA) requires jointly reasoning about perceptual fidelity and prompt alignment, two quality dimensions that are often treated as independent in existing AIGIQA models. However, by re-examining human ratings, we uncover a previously overlooked phenomenon: the two dimensions are interdependent and exhibit both competitive and cooperative interactions during human rating. This observation suggests that a unified model should neither collapse the two dimensions nor rigidly separate them, but rather adaptively negotiate their interplay. Motivated by this insight, we introduce an interaction-aware learning framework that models perception–alignment relations through adversarial and collaborative inference pathways. Instead of designing a rigid dual-branch architecture, our method employs a gated interaction module that dynamically routes features according to the inferred relationship between the two dimensions. Task-aware prompts further modulate the gating behaviour, enabling the model to switch between competition and cooperation when necessary. Experiments across multiple AIGIQA benchmarks demonstrate that our approach not only achieves state-of-the-art accuracy but also yields interpretable interaction patterns, offering a more faithful approximation of human judgment. The codes are available at \url{https://github.com/LQAMEI/ACL-IQA}.
\end{abstract}

\begin{IEEEkeywords}
AI-generated image quality assessment, Adversarial and collaborative learning, 
Mixture-of-Experts.
\end{IEEEkeywords}

\section{Introduction}

\begin{figure}[t]
    \centering
    \includegraphics[width=0.5\textwidth]{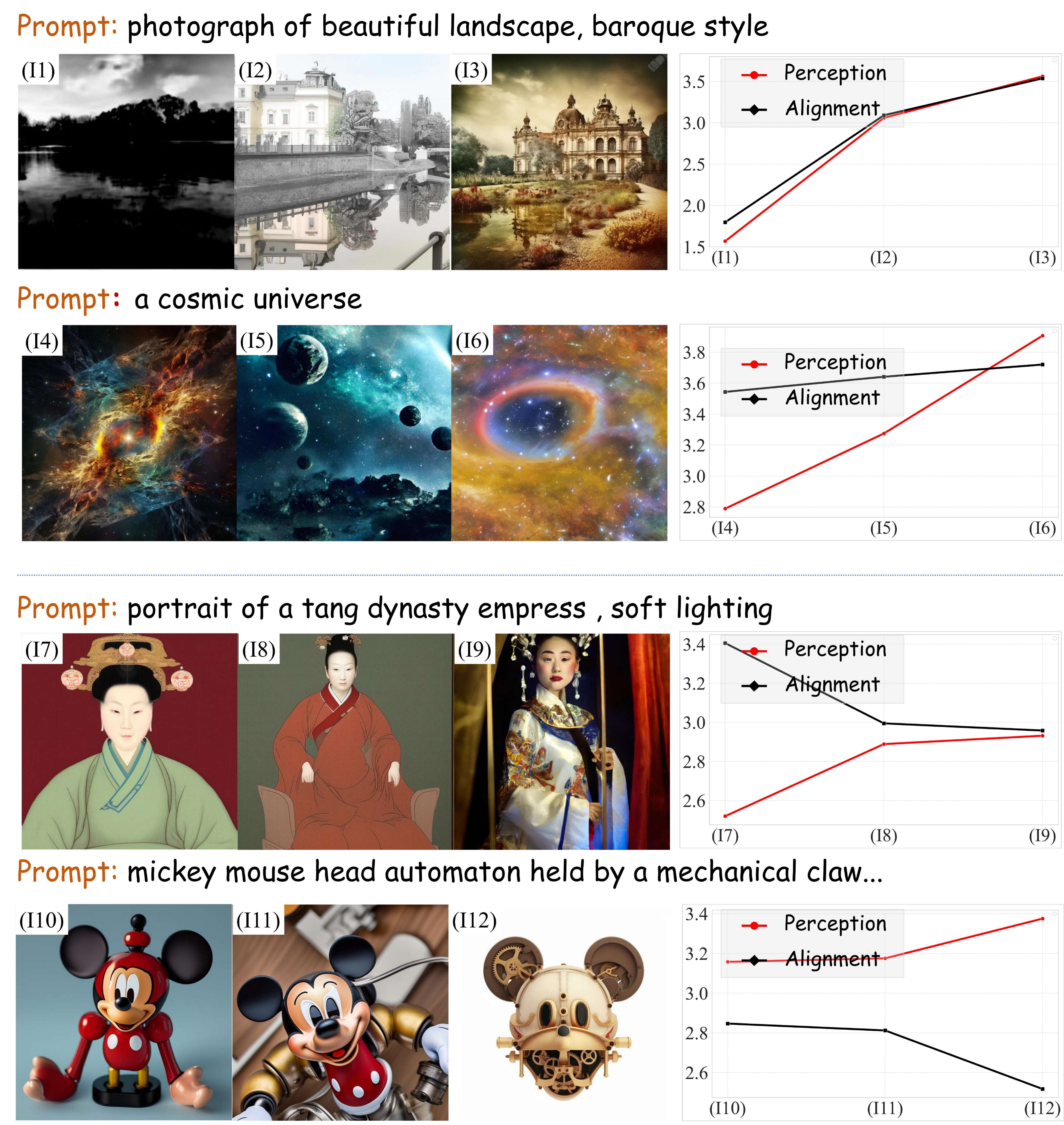}
    \caption{Illustration of adversarial and collaborative interactions in human rating during AIGIQA. The ground-truth quality scores of each dimension are depicted in the line charts.}
    \label{fig:intro}
\end{figure}

Artificial Intelligence Generated Image Quality Assessment (AIGIQA) has emerged as an important problem in evaluating the reliability of modern generative models. Unlike traditional image quality assessment (IQA), where degradations mainly arise from physical limitations (\eg, sensor noise, compression, or enhancement artifacts), the distortions in AIGIs are fundamentally different. They stem from the generative process itself, often guided by user prompts, and can manifest as surreal textures, implausible geometries, or subtle mismatches between image and text semantics. \gr{This paradigm shift transforms AIGIQA into a multi-dimensional evaluation problem that requires assessing both the visual realism of an image (perceptual quality), and its adherence to the textual description (prompt alignment).}

Over the past decades, numerous IQA models \cite{fang2020perceptual, su2020blindly, min2017unified,chen2021learning,chen2025monotonic,chen2026global} have been developed to handle common degradations like noise or blur. However, these methods are inherently limited when confronted with the unique distortions introduced by modern generative models, distortions that are \textit{semantic} rather than \textit{photometric}. Moreover, since they typically \gr{operate solely on visual inputs, they cannot directly} evaluate prompt alignment. To address these limitations, AIGIQA-specific methods have emerged. For example, TIER \cite{yuan2024tier} employs dual encoders for image and text to jointly model their correspondence, while IPCE \cite{peng2024aigc} leverages CLIP \cite{radford2021learning} to capture broader image-text alignment.  
Yet, despite these promising efforts, one crucial aspect remains underexplored: the intrinsic interaction between perceptual and alignment quality. Existing approaches usually treat them as independent tasks, overlooking the subtle ways in which they influence each other during human judgment. As illustrated in Fig.~\ref{fig:intro}, this interaction can be either \textit{collaborative} or \textit{adversarial}. For instance, in cases such as {I1--I3} and {I4--I6}, prompts containing quality-related words cause the two assessments to reinforce each other, where high-quality visuals tend to be perceived as better aligned, and vice versa. In other cases such as {I7--I9} and {I10--I12}, the two aspects behave independently or even conflict, suggesting that they should be disentangled.  
This observation raises an intriguing question:

\begin{quote}
\textit{\textbf{How should we estimate the two types of quality in AIGIs---adversarially, collaboratively, or both?}}
\end{quote}

\noindent In this paper, we argue that the answer is \textbf{\textit{both}}. Human ratings indicate that perceptual quality and prompt alignment do not behave as independent attributes; instead, they exhibit \textit{both adversarial and collaborative interactions}, suggesting that an effective AIGIQA model must explicitly account for these dual modes of behavior. Building on this insight, we design a learning framework that encodes this duality rather than treating it as an incidental property of the architecture. Specifically, we define a collaborative relationship as the synergy where perceptual and alignment dimensions share complementary information. This is evident when textual prompts contain quality related descriptors, where visual realism and semantic adherence mutually reinforce each other. Conversely, an adversarial relationship is defined as the active decoupling of interfering information. This mimics the human cognitive process of suppressing bias, where the model must actively strip away distractor features, such as \gr{aesthetic appeal} that might inflate alignment scores, to ensure independent assessment. 

To operationalize these relationships, we introduce a \textit{Dual-Gated Mixture-of-Experts (DG-MoE)} framework that models perception--alignment interactions through two complementary learning paths. The collaborative path captures mutually reinforcing cues between perceptual and semantic signals to produce joint predictions, while the adversarial path employs a disentangling objective to expose conflicts that may bias quality estimation. DG-MoE utilizes prompt-guided gating functions to selectively route visual features to expert networks specialized for either collaborative reasoning or adversarial disentanglement. By explicitly modeling these interactions, our approach ensures that the network focuses on task specific features without corrupting the shared latent space, thereby capturing a richer set of interaction patterns to produce a more holistic quality estimate.

Our main contributions are summarized as follows:
\begin{itemize}
    \item We provide the first empirical evidence that human AIGI quality ratings exhibit {both adversarial and collaborative interactions} between perceptual and alignment dimensions, motivating a dual-path formulation.
    \item We propose a novel AIGIQA framework that {jointly models adversarial and collaborative interactions}, enabling complementary reasoning about perceptual fidelity and prompt alignment.
    \item We introduce a {Dual-Gated Mixture-of-Experts (DG-MoE)} module that dynamically routes expert features through collaborative or adversarial paths within a unified transformer architecture.
    \item Extensive experiments on multiple AIGIQA benchmarks demonstrate the superiority of our approach and reveal strong complementarity between the two learning paths.
\end{itemize}

\section{Related Work}

\subsection{AIGI Quality Assessment}
The rapid progress of generative models, from Generative Adversarial Networks (GANs)~\cite{goodfellow2020generative} and autoregressive models~\cite{ding2021cogview} to diffusion models~\cite{rombach2022high}, has profoundly accelerated the development of AIGIs. Despite continuous advancements in visual quality, generated images still face challenges such as structural inconsistencies, semantic deviations, and loss of details, preventing AIGIs from fully meeting practical requirements. In response to these challenges, AIGIQA has become a crucial task for optimizing AIGI models~\cite{zhang2023perceptual,li2023agiqa,zhu2025adaptive}. Pioneering studies have released benchmark datasets for subjective evaluation, including AGIQA-1K~\cite{zhang2023perceptual}, AGIQA-3K~\cite{li2023agiqa}, and AIGCIQA2023~\cite{wang2023aigciqa2023}. More recently, contemporary benchmarks have significantly expanded in both scale and diversity to accommodate increasingly complex prompts and generation behaviors. For instance, large-scale databases such as AGIQA-20K~\cite{li2024aigiqa} and EvalMi-50K~\cite{wang2025lmm4lmm} introduce massive human annotations across comprehensive quality dimensions. Concurrently, specialized datasets including EvalMuse-40K~\cite{han2024evalmuse}, GenAI-Bench~\cite{li2024genai}, T2I-CompBench~\cite{huang2023t2i}, GenEval~\cite{ghosh2023geneval}, and RichHF-18K~\cite{liang2024rich} shift the evaluation focus toward subtle compositional errors, fine-grained spatial relations, and localized artifact distributions.

For perceptual quality assessment, classical metrics such as the Inception Score (IS)~\cite{salimans2016improved} and Fréchet Inception Distance (FID)~\cite{heusel2017gans} have been widely adopted to quantify image realism and diversity. Early no-reference IQA methods have provided valuable insights for AIGI quality evaluation. For example, Fang studied perceptual quality of smartphone photography~\cite{fang2020perceptual}, Zhu and Li proposed MetaIQA for adaptive NR-IQA using meta-learning~\cite{zhu2020metaiqa}, and Sun introduced GraphIQA~\cite{sun2022graphiqa}, which models distortion types via a graph representation for blind IQA. Recent studies have advanced this task by incorporating semantic understanding, where hybrid text encoders are employed to enrich feature representations~\cite{fang2024pcqa}, and learnable textual or visual prompts are introduced to enhance prediction accuracy~\cite{fu2024vision}. Concurrently, to better handle diverse structural distortions and generation artifacts, methods such as TReS~\cite{golestaneh2022no}, ManIQA~\cite{yang2022maniqa}, and Re-IQA~\cite{saha2023re} explore multi-scale feature extraction, global transformer backbones, and relative ranking strategies to improve representation robustness.

For alignment assessment, CLIP-based methods~\cite{radford2021learning} have become the dominant paradigm. LIQE~\cite{zhang2023blind} introduces language-guided alignment to map visual features with quality descriptive words. Extending this cross-modal paradigm to AIGIQA, IP-IQA~\cite{qu2024bringing} integrates explicit text prompts to concurrently regress both perceptual and alignment quality, while Peng \etal~\cite{peng2024aigc} propose IPCE to measure the cosine similarity between generated images and tailored prompts. In TIER~\cite{yuan2024tier}, dual encoders for image and text are jointly optimized to regress both perceptual and alignment quality through cross-modal fusion. Yu \etal~\cite{yu2024sf} propose SF-IQA, which integrates quality and alignment scores via a dedicated fusion module. To capture finer-grained correlations, Xia \etal~\cite{xia2024ai} design task-specific prompts to measure both coarse- and fine-grained similarities between AIGIs and their textual descriptions. \gr{To alleviate the entanglement between semantic understanding and distortion perception, MST-CLIPIQA~\cite{meng2026decoupling} adopts two CLIP streams with complementary patch granularities to separately capture global semantic coherence and local artifact-sensitive cues, followed by adaptive cross-scale fusion for quality prediction.} Along this line, the recent wave of Vision-Language Models and Multimodal Large Language Models (MLLMs) has introduced sophisticated evaluators. Models such as FGA-BLIP2~\cite{han2024evalmuse} and RALI~\cite{zhao2025reasoning} focus heavily on fine-grained compositional constraints and human feedback text-matching. Drawing insights from human cognitive mechanisms, Liao \etal~\cite{liao2025plug} proposed the AGQG module, which plugs into existing architectures to accurately calibrate AIGIQA scores. \gr{SC-AGIQA~\cite{li2025text} explicitly addresses semantic misalignment and insufficient fine-grained distortion perception through two complementary modules: an MLLM-assisted semantic alignment module that compares generated image descriptions with the conditioning prompt, and a frequency-domain degradation perception module that enhances sensitivity to subtle perceptual distortions.} Massive MLLM-based architectures such as Q-Align~\cite{wu2024qalign}, MA-AGIQA~\cite{wang2024large}, LMM4LMM~\cite{wang2025lmm4lmm}, DeQA-Score~\cite{you2025teaching}, and Q-Insight~\cite{li2026q} leverage the strong reasoning capabilities of foundation networks to reformulate both perception and alignment assessment as visual question answering or discrete text-defined level generation. While these approaches have achieved remarkable progress, they generally treat perceptual and alignment assessments as independent objectives. The intrinsic interplay between the two, however, remains largely unexplored, leaving a crucial gap in understanding how human perception jointly evaluates quality and alignment in AI-generated content.

\subsection{Mixture of Expert}
Mixture-of-Experts (MoE) models \cite{jacobs1991adaptive} have emerged as a powerful paradigm for enhancing model performance and generalizability without excessively increasing computational complexity. The core idea behind MoE is to introduce multiple expert networks and a gating mechanism that dynamically selects the most relevant experts based on the inputs. In MoE, the gating strategy usually plays a crucial role,  determining both model efficiency and expert selection.
Sparse MoE models \cite{shazeer2017outrageously} improve computational efficiency by utilizing a router to activate only a subset of experts, ensuring that inference time remains comparable to traditional methods while enabling the construction of large-scale models \cite{fedus2022switch}. Task-specific gating networks have been extensively studied to improve expert selection in multi-task settings \cite{kudugunta2021beyond, ma2018modeling, aoki2022heterogeneous, fan2022m3vit}, aiming to select the appropriate parts of a model based on the specific task. For example, Ma \etal propose the MMoE \cite{ma2018modeling} where separate gating networks are designed for each task, enabling selective expert utilization. In M\textsuperscript{3}ViT \cite{fan2022m3vit},  the task-related semantics \gr{are} introduced in the gating design by converting the task information into a one-hot embedding. 
Inspired by these advancements, our work models different experts as distinct convolution kernels and introduces a novel dual gating network towards the adversarial and cooperative learning, respectively.

\section{Methodology}

\begin{figure*}[htp]
    \centering
    \includegraphics[width=1.0\linewidth]{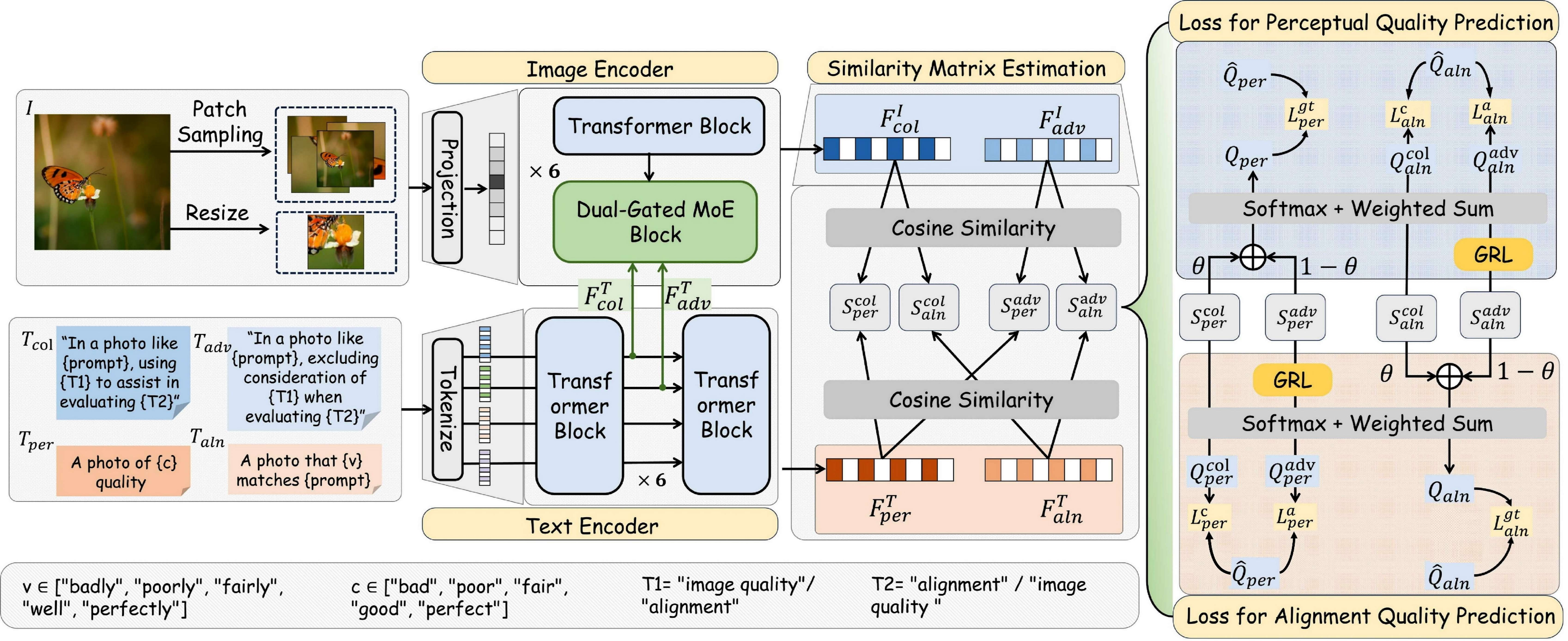}
    \caption{Overview of our AIGIQA model incorporating both adversarial and collaborative learning.
    }
    \label{fig:architecture}
\end{figure*}

\noindent \textbf{Overview.} Given an AIGI and its corresponding prompt, our objective is to predict both its perception quality and alignment quality scores. As depicted in Fig.~\ref{fig:architecture}, our proposed dual-path learning framework comprises three main components: \textit{Multimodal Inputs},  \textit{Text and Image Encoders}, and \textit{Learning Objectives}. Initially, the input AIGI is segmented and resized into multiple patches to effectively capture both local and global information. Additionally, we craft four text templates to instruct the learning (collaborative and adversarial) strategy and quality (perception and alignment) prediction tasks. For the image encoding phase, we introduce a Dual-Gated MoE block, extracting distinct image components tailored for the adversarial and collaborative learning, respectively. This differentiation is achieved through guidance derived from the learning instruction texts. Regarding the learning objectives, we simultaneously predict both perception and alignment quality scores in the collaborative learning, while a Gradient Reversal Layer (GRL) is utilized to disentangle the intertwined features during adversarial learning. The final prediction is obtained by synthesizing the dual learning paths, combining insights from both collaborative and adversarial processes. To facilitate the presentation of our multi-modal dual-path learning framework, we globally standardize the key symbols and definitions in Table~\ref{tab:symbol_table2}. The details of each component are elaborated as follows.

\begin{table}[h]
\caption{Illustration of key mathematical symbols and notations.}
\label{tab:symbol_table2}
\centering
\footnotesize
\setlength{\tabcolsep}{1.8mm}
\begin{tabular}{m{2cm}|m{6cm}} 
\toprule
\multicolumn{1}{c|}{\textbf{Symbol}} & \multicolumn{1}{c}{\textbf{Definition}} \\ 
\midrule
$T_{col}, T_{adv}$ & Textual instructions for collaborative and adversarial learning strategy. \\
\addlinespace[0.5em]
$T_{per}, T_{aln}$ & Textual prompt templates for perception and alignment assessment. \\
\addlinespace[0.5em]
$F_{col}^T, F_{adv}^T$ & Textual learning instruction features derived from templates $T_{col}$ and $T_{adv}$. \\
\addlinespace[0.5em]
$F_{per}^T, F_{aln}^T$ & Textual quality assessment features derived from templates $T_{per}$ and $T_{aln}$. \\
\addlinespace[0.5em]
$F_{col}^I, F_{adv}^I$ & Decoupled image features routed for collaborative and adversarial learning paths. \\
\addlinespace[0.5em]
$G_c(\cdot), G_a(\cdot)$ & Dual gating networks guided by instructional textual features. \\
\addlinespace[0.5em]
$E_j(\cdot), W_j^E$ & The $j$-th difference convolution expert network and its gating weight. \\
\addlinespace[0.5em]

\begin{tabular}[c]{@{}l@{}}$S_{per}^{col}, S_{per}^{adv},$\\ $S_{aln}^{col}, S_{aln}^{adv}$\end{tabular} & Intermediate similarity matrices estimated between decoupled image features and textual quality features. \\
\addlinespace[0.5em]
$S_{per}, S_{aln}$ & Final synthesized similarity matrices aggregated from the collaborative and adversarial paths. \\
\addlinespace[0.5em]
\begin{tabular}[c]{@{}l@{}}$Q_{per}^{col}, Q_{per}^{adv},$\\ $Q_{aln}^{col}, Q_{aln}^{adv}$\end{tabular} & Intermediate quality scores predicted by individual learning paths. \\
\addlinespace[0.5em]
$Q_{per}, Q_{aln}$ & Estimated perception and alignment quality scores predicted by the joint learning framework. \\
\addlinespace[0.5em]
$\hat{Q}_{per}, \hat{Q}_{aln}$ & Ground-truth Mean Opinion Scores (MOS) for human perceptual and alignment quality. \\
\bottomrule
\end{tabular}
\end{table}

\subsection{Multimodal Inputs}
\subsubsection{Image Preprocessing} Given AIGIs of varying sizes, we first segment the image into multiple patches with a standard size of $224 \times 224$. However, segmentation may negatively impact semantic integrity, as a single object can be divided across multiple patches \cite{peng2024aigc}. To address this, our method incorporates both segmented patches and one resized image as input and the final prediction is obtained through a weighted fusion of both types. 

\subsubsection{Text Templates} Two types of text templates are designed in our method. The first set consists of two quality assessment templates designed for different quality level classifications:
\begin{align*}  
T_{per}: & \text{ ``A photo of \{c\} quality''}, \\
           & c \in \{\text{``bad''}, \text{``poor''}, \text{``fair''}, \text{``good''}, \text{``perfect''}\}, \\
T_{aln}: & \text{ ``A photo that \{v\} matches `prompt' ''}, \\
           & v \in \{\text{``badly''}, \text{``poorly''}, \text{``fairly''}, \text{``well''}, \text{``perfectly''}\},  
\end{align*} 
where $T_{per}$ and  $T_{aln}$ correspond to perception quality and alignment quality assessment, respectively. Another two templates are the learning strategy instruction,  serving as guidance for extracting different image components during training:
\begin{align*}  
T_{col}: & \text{ ``In a photo like `prompt', using  \{T1\}  to  assist } \\
          & \text{in evaluating  \{T2\}''}, \\
T_{adv}: & \text{ ``In a photo like `prompt', excluding consideration} \\
          & \text{ of  \{T1\} when evaluating  \{T2\}''}, \\
\text{T1} &= \left\{
\begin{aligned}
\text{``image quality''}  & \text{   if   }   \text{T2 = ``alignment''};\\
\text{``alignment''}  & \text{   if   }     \text{T2 = ``image quality''}.\\
\end{aligned}
\right.
\end{align*} 
Herein, $T_{col}$ and $T_{adv}$ denote the templates correspond to the collaborative and adversarial learning instructions, respectively.

\subsection{Text and Image Encoders}
As shown in Fig.~\ref{fig:architecture}, we design a text encoder and an image encoder to extract the text and image features, respectively. Specifically, the text encoder extracts four types of features: the collaborative and adversarial learning instruction features \(F_{col}^T\) and \(F_{adv}^T\), along with the perception  and alignment quality assessment features \(F_{per}^T\) and \(F_{aln}^T\). The four features  are derived from the text templates  \(T_{col}\), \(T_{adv}\), $T_{per}$ and $T_{aln}$, respectively.  The image encoder extracts two types of image features, namely the collaboratively learned feature \(F_{col}^I\) and the adversarially learned feature \(F_{adv}^I\), representing different components decomposed from the image for the dual-path (collaborative and adversarial) prediction. In particular, we implemented this decomposition through the proposed dual-gated MoE block, where distinct expert networks are designed for specialized image feature extraction. The flow paths of the extracted features are then guided by the learning-instruction features   \(F_{col}^T\) and \(F_{adv}^T\) to facilitate either collaborative or adversarial learning. The architectural details are described as follows.

\subsubsection{Text  Encoder}  
We implement the text encoder following the standard CLIP network. Specifically, a tokenization layer and 12 transformer blocks are adopted. For each block in the odd stages (\(u = 1, 3, \dots, 11\)), we extract the learning instruction features \(F_{col,u}^T\) and \(F_{adv,u}^T\) and input them into the Dual-gated MoE block to tune the gating network.

\begin{figure}
    \centering
    \setlength{\abovecaptionskip}{0.5pt}
    \includegraphics[width=0.8\linewidth]{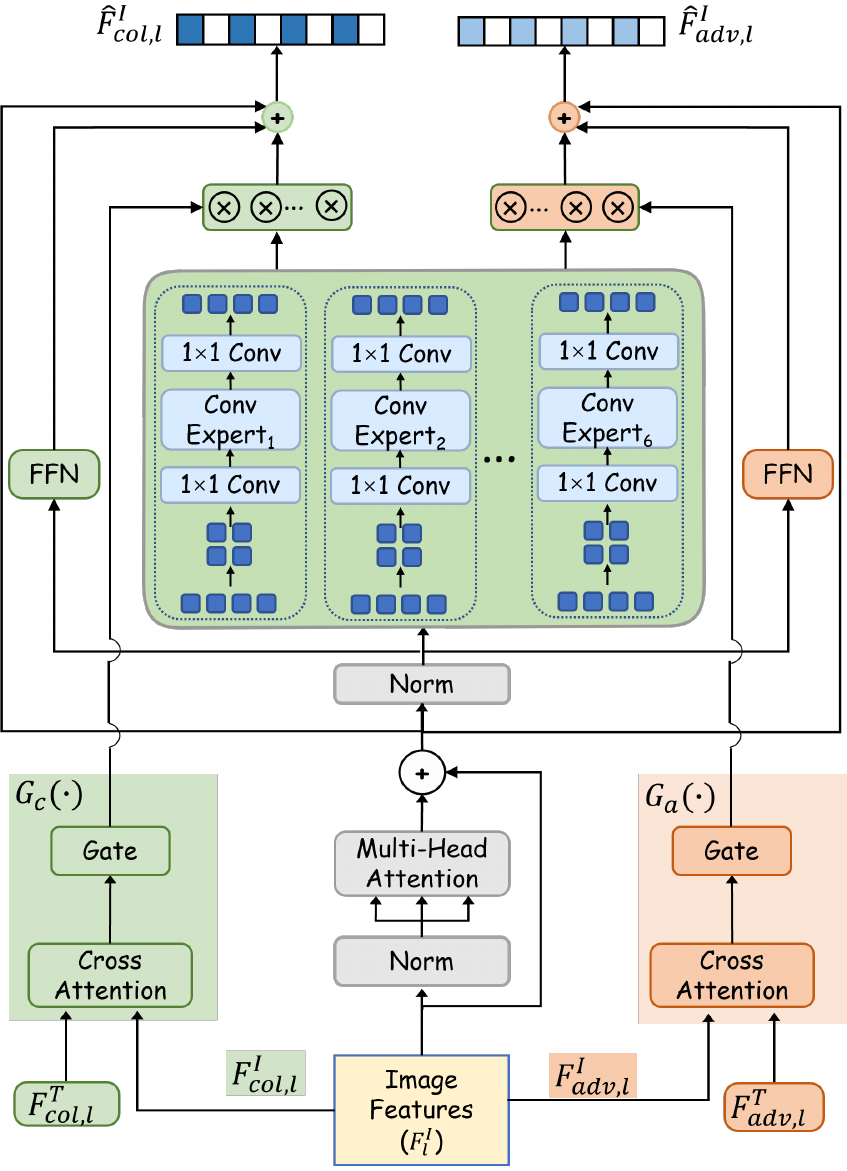}
    \caption{ Design details of our proposed dual-gated MoE block.
    }
    \label{fig:exp}
\end{figure}

\vspace{4mm}
\subsubsection{Image  Encoder} We adopt the Vision Transformer (ViT) \cite{dosovitskiy2020image} as the base architecture for the image encoder. The encoder comprises one linear projection layer, six transformer blocks, and six Dual-gated MoE blocks, with the transformer blocks and Dual-gated MoE blocks arranged alternately. The goal of the image encoder is to extract two types of features, \(F_{col}^I\) and \(F_{adv}^I\), for collaborative and adversarial learning, respectively. As depicted in Fig.~\ref{fig:exp}, we start by initializing the feature extracted from the first transformer block as \(F^I_1\). The outputs of the transformer block at the \(l\)-th stage are denoted as \(F^I_{col,l}\) and \(F^I_{adv,l}\) (\(l = 1, 2, \dots, 6\)). We initialize \(F^I_1 = F^I_{{col},1} = F^I_{{adv},1}\), and the function of the Dual-gated MoE blocks, denoted as \(f(\cdot)\), can be expressed as follows:
\begin{equation}  
\begin{aligned} 
\hat{F}^I_{{col},l}, \hat{F}^I_{{adv},l} & = f(F^I_{col, l}, F^I_{adv, l}) \\
             & =  f_c(F^I_{col, l}),   f_a(F^I_{adv, l}),
\end{aligned}  
\end{equation}  
where \( f_c(\cdot) \) and \( f_a(\cdot) \) represent the two sub-networks responsible for collaborative and adversarial feature extraction, respectively. To simplify the explanation, we take \( F^I_{col, l} \) as an example to illustrate the design details of \( f_c(\cdot) \), as both \( f_c(\cdot) \) and \( f_a(\cdot) \) follow the same design philosophy, as shown in Fig.~\ref{fig:exp}:
\begin{equation}  
\begin{aligned} 
\hat{F}^I_{col,l} & = f_c(F^I_{col,l}) \\
             & = F^{att}_{{col,l}} + \text{FFN}(\text{LN}(F^{att}_{{col,l}})) + F^{moe}_{{col,l}},
\end{aligned}  
\end{equation}  
with
\begin{equation} 
 F^{att}_{{col,l}} = F^I_{col,l} + \text{MSA}(\text{LN}(F^I_{col,l})),
\end{equation} 
\begin{equation} 
 F^{moe}_{{col,l}} = \sum_{j=1}^{N_e} W^E_j E_j (\text{LN}(F^{att}_{{col,l}})),
 \label{eq:moe}
\end{equation} 
where LN, FFN, and MSA represent the standard layer normalization, feed-forward network, and multi-head self-attention layers in the  ViT. \( N_e \) denotes the number of experts in the Dual-gated MoE block. \( E_j(\cdot) \) represents the function of the \( j \)-th expert network, and \( W^E_j \) is the weight predicted by a gating network \( G_c(\cdot) \).  The design details of the \( G_c(\cdot) \) and \( E_j(\cdot) \) are elaborated as follows.

\begin{itemize}
\vspace{2mm}
\item \textbf{Dual Gating Network $G(\cdot)$.} The dual gating network aims to estimate the weight \( W^E_j \) of each expert output with the two learning text features \( F_{col,u}^T \) and $ F_{adv,u}^T \mid u=(2l-1) $  as instruction. We denote the gating networks guided by \( F_{c,u}^T \) and \( F_{a,u}^T \) as \( G_c(\cdot) \) and \( G_a(\cdot) \), respectively. As shown in Fig.~\ref{fig:exp}, \( G_c(\cdot) \) and \( G_a(\cdot) \) share the same design. For simplicity, we use \( G_c(\cdot) \) as an example then the \( W^E_j \) thus  can be calculated by:
\begin{equation} 
\begin{aligned} 
W^E_j &= G_c( F^I_{col,l}, F^T_{col,2l-1})  \\
      &= \text{Softmax}(\text{Topk}(W^{gate}_jF_{gate}, k)),
\end{aligned}  
\end{equation}
with
\begin{equation} 
F_{gate} = \text{AvgPool}(\text{CrossAtten}(F^I_{col,l}, F^T_{col,u})),
\end{equation}
where \( \text{Softmax}(\cdot) \) and \( \text{Topk}(\cdot) \) represent the functions of softmax activation and top-k selection, respectively. \( W^{gate}_j \) are learnable weights that project the gating feature \( F_{gate} \) to a single value, which is used to weight the \( j \)-th expert's output. The \( \text{CrossAtten}(\cdot) \) is defined as follows:
\begin{equation}  
\begin{aligned}  
  \text{CrossAtten}(F^I_{col,l}, F^T_{col,u}) = \text{Softmax}\left( \frac{QK^T}{\sqrt{D}} \right) V,\\
  Q =F^I_{col,l} W^Q, \quad K =F^T_{col,u} W^K, \quad V =F^T_{col,u} W^V, 
\end{aligned}  
\end{equation}  
where $W^K$, $W^V$,  and $W^Q$ are learnable projection matrices. $D$ represents the dimension of the keys.

\vspace{5mm}
\item \textbf{Expert Network  $E_j(\cdot)$.} We design the expert networks using difference convolution layers \cite{su2021pixel, kong2024moe} (denoted as \( \text{ConvExpert} \)\gr{)} in Fig.~\ref{fig:exp}. The design philosophy lies in that the gradient information between adjacent tokens, captured by the difference convolution kernels, provides more fine-grained and high-frequency details. This plays a complementary role to the transformer network, which is more inclined to capture low-frequency information \cite{park2022how}. In our method, we adopt a total of six types of ConvExperts, including Vanilla convolution and five difference convolutions: Horizontal Difference Convolution (HDC), Vertical Difference Convolution (VDC), Central Difference Convolution (CDC), Angular Difference Convolution (ADC), and Radial Difference Convolution (RDC). The kernels of the five difference convolutions are depicted in Fig.~\ref{fig:diff_conv}. More specifically, denoting the input feature \( \text{LN}(F^{att}F^I_{col,l}) \) (see Eqn.~(\ref{eq:moe})) as \( F_{\exp} \), the ConvExpert with the \( j \)-th kernel can be expressed as follows:
\begin{equation}  
\begin{aligned}  
E(F_{exp})_j &= \text{Conv}_{{up}} \left( \text{Conv}_j \left( \text{Conv}_{{down}} (F_{exp}) \right) \right),
\end{aligned}  
\end{equation}
where the $\text{Conv}_j$ means the convolutional layer with the $j$-th kernel. $\text{Conv}_{{down}}$ and $\text{Conv}_{{up}}$ are two $1\times1$ convolutional layers aiming for the channel dimension downsampling and upsampling (restoring) for computational efficiency.
\end{itemize}

\noindent Through the expert networks, different components can be extracted from the image and routed to the collaborative and adversarial learning paths with the guidance obtained by \( G_c(\cdot) \) and \( G_a(\cdot) \), respectively.

\subsection{Collaborative and Adversarial Learning} 
Based on the obtained image features \( F_{col}^I \), \( F_{adv}^I \) and quality assessment features \( F_{per}^T \) and \( F_{aln}^T \), we compute the cosine similarity between them, resulting in four types of similarity matrices:
\begin{equation}  
\begin{aligned}  
S_{per}^{col} &= \frac{F_{col}^I \odot F_{per}^T}{\|F_{col}^I\| \cdot \| F_{per}^T\|},  \quad S_{per}^{adv} = \text{GRL}(\frac{F_{adv}^I \odot F_{per}^T}{\|F_{adv}^I\| \cdot \| F_{per}^T\|}), \\
S_{aln}^{col} &= \frac{F_{col}^I \odot F_{aln}^T}{\|F_{col}^I\| \cdot \| F_{aln}^T\|},   \quad  S_{aln}^{adv} = \text{GRL}(\frac{F_{adv}^I \odot F_{aln}^T}{\|F_{adv}^I\| \cdot \| F_{aln}^T\|}) ,
\end{aligned}  
\end{equation}
where \( \text{GRL}(\cdot) \) denotes the Gradient Reversal Layer \cite{ganin2015unsupervised}, which is designed to ablate the specific quality information through negative gradient backpropagation. Additionally, we employ a trainable weighting scheme for the fusion of \( S_{per}^{col} \) and \( S_{per}^{adv} \), which serves as the final estimation for the perception and alignment quality:
\begin{equation}  
\begin{aligned}  
S_{per} &= S_{per}^{col} \times \theta + S_{per}^{adv} \times (1-\theta), \\
S_{aln} &= S_{aln}^{col} \times \theta + S_{aln}^{adv} \times (1-\theta),
\end{aligned}  
\end{equation}
where $\theta$ is a trainable parameter. Herein, the matrices \( S_{per}^{col}, S_{aln}^{col}, S_{per}^{adv}, S_{aln}^{adv}, S_{per}, S_{aln} \in \mathbb{R}^{M \times 5} \), where \( M \) represents the number of image patches, and \text{``5"} denotes the five quality levels defined in \( T_{per} \) and \( T_{aln} \). We denote the \( M \)-th patch as the resized one. 
Then, we convert the six similarity matrices into their corresponding quality scores \( Q_{per}^{col}, Q_{aln}^{col}, Q_{per}^{adv}, Q_{aln}^{adv}, Q_{per}, Q_{aln} \), respectively. Taking the conversion from \( S_{per}^{col} \) to \( Q_{per}^{col} \) as an example:
\begin{equation}  
Q_{per}^{col} =  \frac{5}{4} \left( \sum_{i=1}^5 i \times P_i - 1 \right),  
\end{equation}
with
\begin{equation}
\begin{aligned}
P_{i,j} &= \text{Softmax}(S_{per}^{col}),  \\
P_{i} &=\left( \frac{\sum_{j=1}^{M-1} P_{i,j}}{M-1} + P_{i, M} \right) / 2,
\end{aligned}
\end{equation}
where the \(\text{Softmax}\) function is applied along the quality level dimension. $P_{i,j}$ denotes the quality probability of the $j$-th patch under the $i$-th level. Finally, we define the supervision loss functions as follows:

\begin{figure}[t]
    \centering
    \setlength{\abovecaptionskip}{0.5pt}
    \includegraphics[width=1\linewidth]{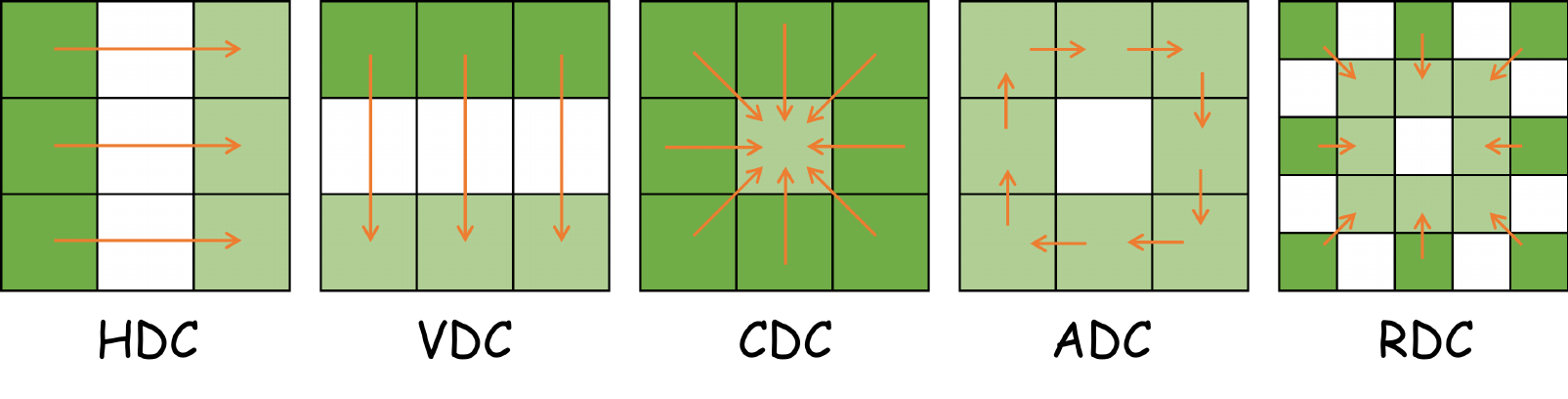}
    \caption{Illustration of five types of convolution kernels~\cite{kong2024moe}. The arrow represents the difference between the weight at its starting point and the weight at its endpoint.
    }
    \label{fig:diff_conv}
\end{figure}
\begin{itemize}
\item For the perception quality assessment task:
\begin{equation}  \label{eqall1}
L_{per} =  L^{gt}_{per}+ \lambda_1L^{c}_{aln} +\lambda_2 L^{a}_{aln} + \lambda_3 L_{{moe}},
\end{equation}
where $\lambda_1$, $\lambda_2$, and $\lambda_3$ are three hyperparameters. $L_{{moe}}$ is an expert utilization balancing loss that encourages all experts to have equal importance \cite{shazeer2017outrageously}, and 
\begin{equation}  
\begin{aligned} 
L^{gt}_{per} &=  |\hat{Q}_{per} - Q_{per}|, \\
L^{c}_{aln} &=  |\hat{Q}_{aln} - Q^c_{aln}|, \\
L^{a}_{aln} &=  |\hat{Q}_{aln} - Q^a_{aln}|, 
\end{aligned} 
\end{equation}
where $\hat{Q}_{per}$ and $\hat{Q}_{aln}$ are the ground-truth perception and alignment quality scores, respectively.

\item For the alignment quality assessment task:
\begin{equation}  \label{eqall2}
L_{aln} =  L^{gt}_{aln} + \lambda_1 L^{c}_{per} + \lambda_2 L^{a}_{per} + \lambda_3  L_{moe},
\end{equation}
where 
\begin{equation}  
\begin{aligned} 
L^{gt}_{{aln}} &=  |\hat{Q}_{aln} - Q_{aln}|, \\
L^{c}_{{per}} &=  |\hat{Q}_{per} - Q^c_{per}|, \\
L^{a}_{{per}} &=  |\hat{Q}_{per} - Q^a_{per}|. 
\end{aligned} 
\end{equation}

\end{itemize}
\noindent In the testing phase, we adopt $Q_{{per}}$ and $Q_{{aln}}$ as the estimated perception quality score and alignment quality score, respectively.

\section{Experiment}

\subsection{Experimental Setting} 
\subsubsection{Datasets} We conduct comprehensive experiments on five publicly available AIGIQA datasets: AGIQA-1K~\cite{zhang2023perceptual}, AGIQA-3K~\cite{li2023agiqa}, AIGCIQA2023~\cite{wang2023aigciqa2023}, and two massive benchmarks, AGIQA-20K~\cite{li2024aigiqa} and EvalMi-50K~\cite{wang2025lmm4lmm}. 
AGIQA-1K consists of 1,080 images generated by Stable Inpainting v1 and Stable Diffusion v2, using 180 prompts derived from high-frequency keywords. Each image is assigned a Mean Opinion Score (MOS) based on both perception quality and text-image correspondence. AGIQA-3K contains 2,982 images generated by six text-to-image (T2I) models. Each image is annotated with both perception quality and text-image alignment score. AIGCIQA2023 includes 2,400 images generated by six T2I models using 100 prompts. The annotations cover perception quality, authenticity, and correspondence. To rigorously validate scalability against modern generative priors, AGIQA-20K comprises 20,000 images generated by 15 mainstream models across diverse sampling configurations. Furthermore, EvalMi-50K features 50,400 images synthesized by 24 recent text-to-image generators (e.g., DALL-E 3, Flux, Stable Diffusion 3.5) with highly complex textual prompts, providing robust MOS for Perception and Correspondence. Evaluations on fine-grained alignment and artifact datasets (EvalMuse-40K, GenAIBench, and RichHF-18K) are provided in the Supplementary Material.

\subsubsection{Implementation Details} We use the CLIP model with the ViT-B/32 architecture as our base network. For single-dimension evaluations including AGIQA-1K and the authenticity assessments of AIGCIQA2023 and AGIQA-20K, we retain only the single-gated MoE from our original model. The experiments are trained for 200 epochs on the early-stage datasets and for 50 epochs when evaluated on the massive contemporary benchmarks (AGIQA-20K and EvalMi-50K). During training, the batch size is set to 16, and we use the Adam optimizer with an initial learning rate of $5 \times 10^{-6}$ and a weight decay of $1 \times 10^{-3}$. In the dual gating network, we select the top-3 expert outputs for learning, employing the Noisy Top-K Gating strategy~\cite{shazeer2017outrageously} during MoE network training. We set $\lambda_1 = \lambda_2 = 0.5$ and $\lambda_3 = 1.0$ in Eqn.~(\ref{eqall1}) and Eqn.~(\ref{eqall2}). The Spearman Rank Correlation Coefficient (SRCC) and the Pearson Linear Correlation Coefficient (PLCC) are adopted as our evaluation metrics. To guarantee reproducibility and ensure fairness in comparisons, we strictly follow standardized data-splitting protocols: for AGIQA-20K, we directly utilize the train/test split provided by the dataset repository; for EvalMi-50K, we follow the 5-fold cross-validation protocol established in the LMM4LMM~\cite{wang2025lmm4lmm}; and for all other datasets, the experiments are repeated 10 times under identical data splits, with the average results reported.

\subsection{Experimental Results}
We denote our Adversarial and Collaborative  Learning based IQA model as ACL-IQA.

\subsubsection{Prediction Ability Comparison}
\begin{table}[t]
\caption{Performance comparison of perception quality prediction on the AGIQA-1K. The best two results are in boldface.}
\centering
\setlength{\tabcolsep}{2mm}

\begin{tabular}{l|ccc}
\midrule
{Method} & SRCC & PLCC & Mean Score\\
\cmidrule{1-4} 
CNNIQA \cite{kang2014convolutional} &0.5800 &0.7139 &0.6470\\
DBCNN \cite{zhang2018blind} &0.7491 &0.8211 &0.7851 \\
MGQA \cite{wang2021multi} &0.6011 &0.6760 &0.6386\\
CLIP-IQA \cite{wang2023exploring} &0.8227 &0.8411 &0.8319\\
StairIQA \cite{sun2023blind} &0.5504 &0.6088 &0.5796\\
PSCR \cite{yuan2023pscr}  & 0.8430 &0.8403 &0.8417\\
TIER \cite{yuan2024tier}  &0.8266 &0.8297 &0.8282\\
IP-IQA \cite{qu2024bringing}  &0.8401 & \textbf{0.8922} & 0.8662\\
IPCE  \cite{peng2024aigc} &\textbf{0.8535} & 0.8792 &\textbf{0.8664}\\

\cmidrule{1-4} 
\rowcolor{green!8!white}
ACL-IQA (Ours)   & \textbf{0.8617}& \textbf{0.8885} & \textbf{0.8751} \\
\midrule
\end{tabular}
\label{tab:agiqa1k-p}
\end{table}

\begin{table*}[htbp]
\caption{Performance comparison of perception and alignment quality prediction on the AGIQA-3K dataset. The best two results in each column are highlighted in boldface. * denotes methods evaluated by directly loading pretrained model weights for inference.}
\centering
\setlength{\tabcolsep}{1.5mm}
\begin{tabular}{l|ccc|l|ccc}
\toprule
\multicolumn{4}{c|}{\textbf{Perception Quality}} & \multicolumn{4}{c}{\textbf{Alignment Quality}} \\
\cmidrule(lr){1-4} \cmidrule(lr){5-8}
\textbf{Method} & SRCC & PLCC & Mean Score & \textbf{Method} & SRCC & PLCC & Mean Score \\
\midrule 
CNNIQA~\cite{kang2014convolutional} & 0.7478 & 0.8469 & 0.7974 & CLIP~\cite{radford2021learning} & 0.5972 & 0.6839 & 0.6406 \\
DBCNN~\cite{zhang2018blind} & 0.8207 & 0.8759 & 0.8483 & ImageReward~\cite{xu2023imagereward} & 0.7298 & 0.7862 & 0.7580 \\
CLIP-IQA~\cite{wang2023exploring} & 0.8426 & 0.8053 & 0.8240 & HPS~\cite{wu2023human} & 0.6623 & 0.7008 & 0.6816 \\
ManIQA~\cite{yang2022maniqa} & 0.8767 & 0.9105 & 0.8936 & ManIQA~\cite{yang2022maniqa} & -- & -- & -- \\
Re-IQA~\cite{saha2023re} & 0.8424 & 0.8890 & 0.8657 & Re-IQA~\cite{saha2023re} & -- & -- & -- \\
TReS~\cite{golestaneh2022no} & 0.8297 & 0.8832 & 0.8565 & TReS~\cite{golestaneh2022no} & -- & -- & -- \\
PSCR~\cite{yuan2023pscr} & 0.8498 & 0.9059 & 0.8779 & PickScore~\cite{kirstain2023pick} & 0.7320 & 0.7791 & 0.7556 \\
CLIP-AGIQA~\cite{tang2024clip} & 0.8618 & 0.8978 & 0.8798 & StairReward~\cite{li2023agiqa} & 0.7472 & 0.8529 & 0.8001 \\
LIQE~\cite{zhang2023blind} & 0.8742 & 0.8934 & 0.8838 & LIQE~\cite{zhang2023blind} & 0.7338 & 0.8205 & 0.7772 \\
TIER~\cite{yuan2024tier} & 0.8251 & 0.8821 & 0.8536 & TIER~\cite{yuan2024tier} & 0.6495 & 0.7988 & 0.7242 \\
IP-IQA~\cite{qu2024bringing} & 0.8634 & 0.9116 & 0.8875 & IP-IQA~\cite{qu2024bringing} & 0.7578 & 0.8544 & 0.8061 \\
IPCE~\cite{peng2024aigc} & 0.8841 & \textbf{0.9246} & \textbf{0.9044} & IPCE~\cite{peng2024aigc} & 0.7697 & 0.8725 & 0.8211 \\
Q-Align~\cite{wu2024qalign} & 0.8058 & 0.8268 & 0.8163 & Q-Align~\cite{wu2024qalign} & 0.7820 & 0.8477 & 0.8149 \\
MA-AGIQA~\cite{wang2024large} & \textbf{0.8872} & 0.9126 & 0.8999 & MA-AGIQA~\cite{wang2024large} & -- & -- & -- \\
FGA-BLIP2~\cite{han2024evalmuse} & 0.8659 & 0.9121 & 0.8890 & FGA-BLIP2~\cite{han2024evalmuse} & 0.7089 & 0.8209 & 0.7649 \\
RALI*~\cite{zhao2025reasoning} & 0.7149 & 0.7785 & 0.7467 & RALI*~\cite{zhao2025reasoning} & -- & -- & -- \\
Q-Insight*~\cite{li2026q} & 0.7649 & 0.8086 & 0.7868 & Q-Insight*~\cite{li2026q} & 0.7250 & 0.7888 & 0.7569 \\
DeQA-Score*~\cite{you2025teaching} & 0.7544 & 0.8066 & 0.7805 & DeQA-Score*~\cite{you2025teaching} & 0.4718 & 0.5818 & 0.5268 \\
LMM4LMM~\cite{wang2025lmm4lmm} & 0.8740 & 0.9161 & 0.8951 & LMM4LMM~\cite{wang2025lmm4lmm} & \textbf{0.8298} & \textbf{0.8976} & \textbf{0.8637} \\
\midrule 
\rowcolor{green!8!white}
\textbf{ACL-IQA (Ours)} & \textbf{0.8969} & \textbf{0.9282} & \textbf{0.9126} & \textbf{ACL-IQA (Ours)} & \textbf{0.7832} & \textbf{0.8800} & \textbf{0.8316} \\
\bottomrule
\end{tabular}
\label{tab:agiqa3k-dual}
\end{table*}

\begin{table*}[htbp]
\caption{Performance comparison of perception, authenticity, and alignment quality prediction on the AIGCIQA2023 dataset. The best two results in each column are highlighted in boldface. * denotes methods evaluated by directly loading pretrained model weights for inference.}
\centering
\setlength{\tabcolsep}{1.1mm}
\begin{tabular}{l|ccc|ccc|ccc}
\toprule
\multirow{2}{*}{\textbf{Method}} & \multicolumn{3}{c|}{\textbf{Perception}} & \multicolumn{3}{c|}{\textbf{Authenticity}} & \multicolumn{3}{c}{\textbf{Alignment}} \\
\cmidrule(lr){2-4} \cmidrule(lr){5-7} \cmidrule(lr){8-10}
& SRCC & PLCC & Mean Score & SRCC & PLCC & Mean Score & SRCC & PLCC & Mean Score \\
\midrule 
CNNIQA~\cite{kang2014convolutional} & 0.7160 & 0.7937 & 0.7549 & 0.5958 & 0.5734 & 0.5846 & 0.4758 & 0.4937 & 0.4848 \\
ResNet34~\cite{he2016deep} & 0.7229 & 0.7578 & 0.7404 & 0.5998 & 0.6285 & 0.6142 & 0.7058 & 0.7153 & 0.7106 \\
TReS~\cite{golestaneh2022no} & 0.8064 & 0.8347 & 0.8206 & -- & -- & -- & -- & -- & -- \\
ManIQA~\cite{yang2022maniqa} & 0.8269 & 0.8464 & 0.8367 & -- & -- & -- & -- & -- & -- \\
Re-IQA~\cite{saha2023re} & 0.8372 & 0.8721 & 0.8547 & -- & -- & -- & -- & -- & -- \\
LIQE~\cite{zhang2023blind} & 0.8534 & 0.8758 & 0.8646 & -- & -- & -- & 0.7780 & 0.7696 & 0.7738 \\
CLIP-IQA~\cite{wang2023exploring} & 0.7802 & 0.8140 & 0.7971 & 0.6726 & 0.6627 & 0.6677 & 0.5799 & 0.5702 & 0.5751 \\
PSCR~\cite{yuan2023pscr} & 0.8371 & 0.8588 & 0.8480 & 0.7828 & 0.7750 & 0.7789 & 0.7465 & 0.7397 & 0.7431 \\
IPCE~\cite{peng2024aigc} & \textbf{0.8640} & \textbf{0.8788} & \textbf{0.8714} & \textbf{0.8097} & \textbf{0.7998} & \textbf{0.8048} & 0.7979 & 0.7887 & 0.7933 \\
CLIP-AGIQA~\cite{tang2024clip} & 0.8140 & 0.8302 & 0.8221 & 0.7940 & 0.7797 & 0.7869 & 0.7036 & 0.6994 & 0.7015 \\
Q-Align~\cite{wu2024qalign} & 0.7975 & 0.8300 & 0.8138 & -- & -- & -- & 0.7398 & 0.7157 & 0.7278 \\
MA-AGIQA~\cite{wang2024large} & 0.8384 & 0.8606 & 0.8495 & -- & -- & -- & -- & -- & -- \\
TIER~\cite{yuan2024tier} & 0.8326 & 0.8526 & 0.8426 & 0.7485 & 0.7403 & 0.7444 & 0.7210 & 0.7148 & 0.7179 \\
FGA-BLIP2~\cite{han2024evalmuse} & 0.8326 & 0.8691 & 0.8509 & -- & -- & -- & 0.7820 & 0.7765 & 0.7793 \\
RALI*~\cite{zhao2025reasoning} & 0.7772 & 0.7964 & 0.7868 & -- & -- & -- & -- & -- & -- \\
Q-Insight*~\cite{li2026q} & 0.8027 & 0.7975 & 0.8001 & -- & -- & -- & 0.6504 & 0.6269 & 0.6387 \\
DeQA-Score*~\cite{you2025teaching} & 0.8024 & 0.8304 & 0.8164 & -- & -- & -- & 0.4232 & 0.4503 & 0.4368 \\
LMM4LMM~\cite{wang2025lmm4lmm} & 0.8344 & 0.8632 & 0.8488 & -- & -- & -- & \textbf{0.8208} & \textbf{0.8141} & \textbf{0.8175} \\
\midrule 
\rowcolor{green!8!white}
\textbf{ACL-IQA (Ours)} & \textbf{0.8720} & \textbf{0.8865} & \textbf{0.8793} & \textbf{0.8180} & \textbf{0.8083} & \textbf{0.8132} & \textbf{0.8092} & \textbf{0.8005} & \textbf{0.8049} \\
\bottomrule
\end{tabular}
\label{tab:agiqa2023pac2}
\end{table*}

\begin{table}[htbp]
\caption{Performance comparison on the AGIQA-20K dataset. The best two results in each column are highlighted in boldface. * denotes methods evaluated by directly loading pretrained model weights for inference.}
\label{tab:agiqa20k}
\centering
\setlength{\tabcolsep}{2mm}
\begin{tabular}{l|ccc}
\toprule
\textbf{Method} & \textbf{SRCC} & \textbf{PLCC} & \textbf{Mean Score} \\
\midrule
CNNIQA~\cite{kang2014convolutional} & 0.5968 & 0.5913 & 0.5941 \\
DBCNN~\cite{zhang2018blind} & 0.8506 & 0.8688 & 0.8597 \\
HyperIQA~\cite{su2020blindly} & 0.8160 & 0.8320 & 0.8240 \\
MUSIQ~\cite{ke2021musiq} & 0.8320 & 0.8640 & 0.8480 \\
ManIQA~\cite{yang2022maniqa} & 0.8500 & 0.8870 & 0.8685 \\
StairIQA~\cite{sun2023blind} & 0.7890 & 0.8420 & 0.8155 \\
LIQE~\cite{zhang2023blind} & 0.8620 & 0.7980 & 0.8300 \\
Q-Align~\cite{wu2024qalign} & \textbf{0.8740} & 0.8890 & 0.8815 \\
MA-AGIQA~\cite{wang2024large} & 0.8640 & \textbf{0.9050} & \textbf{0.8845} \\
DeQA-Score*~\cite{you2025teaching} & 0.4779 & 0.5535 & 0.5157 \\
Q-Insight*~\cite{li2026q} & 0.5359 & 0.6172 & 0.5766 \\
LMM4LMM*~\cite{wang2025lmm4lmm} & 0.7284 & 0.6638 & 0.6961 \\
\midrule
\rowcolor{green!8!white}
\textbf{ACL-IQA (Ours)} & \textbf{0.8858} & \textbf{0.9113} & \textbf{0.8986} \\
\bottomrule
\end{tabular}
\vspace{-2em}
\end{table}

\begin{table*}[htbp]
\caption{Performance comparison on the EvalMi-50K dataset across perception and alignment dimensions. The best two results in each column are highlighted in boldface.}
\centering
\setlength{\tabcolsep}{2mm}
\begin{tabular}{l|ccc|ccc}
\toprule
\multirow{2}{*}{\textbf{Method}} & \multicolumn{3}{c|}{\textbf{Perception Quality}} & \multicolumn{3}{c}{\textbf{Alignment Quality}} \\
\cmidrule(lr){2-4} \cmidrule(lr){5-7}
& SRCC & PLCC & Mean Score & SRCC & PLCC & Mean Score \\
\midrule
CNNIQA~\cite{kang2014convolutional} & 0.4348 & 0.5583 & 0.4966 & 0.1186 & 0.0791 & 0.0989 \\
DBCNN~\cite{zhang2018blind} & 0.5525 & 0.6181 & 0.5853 & 0.3301 & 0.3515 & 0.3408 \\
HyperIQA~\cite{su2020blindly} & 0.5872 & 0.6768 & 0.6320 & 0.5348 & 0.5447 & 0.5398 \\
TReS~\cite{golestaneh2022no} & 0.3935 & 0.4301 & 0.4118 & 0.1406 & 0.1520 & 0.1463 \\
MUSIQ~\cite{ke2021musiq} & 0.7985 & 0.8379 & 0.8182 & 0.5310 & 0.5510 & 0.5410 \\
StairIQA~\cite{sun2023blind} & 0.8268 & 0.8645 & 0.8457 & 0.5890 & 0.6089 & 0.5990 \\
LIQE~\cite{zhang2023blind} & 0.8106 & 0.8268 & 0.8187 & 0.5617 & 0.5777 & 0.5697 \\
Q-Align~\cite{wu2024qalign} & 0.8311 & 0.8505 & 0.8408 & 0.4754 & 0.4918 & 0.4836 \\
DeepSeekVL2(1B)~\cite{wu2024deepseek} & 0.7899 & 0.8253 & 0.8076 & 0.7817 & 0.7991 & 0.7904 \\
Qwen2.5-VL(8B)~\cite{wang2024qwen2} & 0.6990 & 0.7495 & 0.7243 & 0.8008 & 0.8219 & 0.8114 \\
Llama3.2-Vision(11B)~\cite{meta2024llama} & 0.7555 & 0.7891 & 0.7723 & 0.6403 & 0.6461 & 0.6432 \\
FGA-BLIP2~\cite{han2024evalmuse} & \textbf{0.8715} & \textbf{0.8969} & \textbf{0.8842} & \textbf{0.8121} & \textbf{0.8309} & \textbf{0.8215} \\
\midrule 
\rowcolor{green!8!white}
\textbf{ACL-IQA (Ours)} & \textbf{0.8810} & \textbf{0.9050} & \textbf{0.8930} & \textbf{0.8664} & \textbf{0.8795} & \textbf{0.8730} \\
\bottomrule
\end{tabular}
\label{tab:evalmi}
\end{table*}

To evaluate the effectiveness of our model, we conduct comprehensive comparisons with existing IQA and AIGIQA methods across five benchmarks. The results on early-stage datasets, summarized in Tables~\ref{tab:agiqa1k-p} to \ref{tab:agiqa2023pac2}, demonstrate that our approach consistently achieves highly competitive or state-of-the-art performance. On the AGIQA-1K dataset, the integration of a dual-gated mixture-of-experts (MoE) module and a semantic-guided gating mechanism enables our model to effectively capture text-relevant quality cues, outperforming previous methods with clear gains in correlation with human ratings. For AGIQA-3K, our method exhibits superior performance in perception quality by surpassing MA-AGIQA, and remains highly competitive in alignment quality, trailing only LMM4LMM, reflecting its robustness in modeling complex cross-modal dependencies. Moreover, on the challenging AIGCIQA2023 benchmark, our model delivers the best results in Perception and Authenticity, and secures the second-best performance in Alignment, showing strong generalization and balanced improvement across diverse quality aspects.

To thoroughly demonstrate the scalability of our model, we extend our evaluation to massive contemporary datasets, specifically AGIQA-20K and EvalMi-50K. Faced with extreme prompt complexity and diverse generative artifacts, Table~\ref{tab:agiqa20k} shows that ACL-IQA establishes a new state-of-the-art on the large-scale AGIQA-20K dataset, outperforming strong multimodal evaluators like MA-AGIQA and Q-Align. Similarly, on EvalMi-50K as detailed in Table~\ref{tab:evalmi}, our framework consistently surpasses recent strong vision-language models such as FGA-BLIP2, Qwen2.5-VL (8B), and Llama3.2-Vision (11B) across both perception and alignment dimensions. Its ability to deliver such robust evaluations without relying on massive parameter scales highlights its core architectural superiority.

These results confirm the effectiveness of our approach in capturing both perceptual and alignment quality. By leveraging the global modeling capability of the encoder alongside local feature extraction from convolutional experts, our model effectively balances coarse- and fine-grained feature representations. Furthermore, the proposed adversarial and collaborative learning mechanism successfully disentangles perception and alignment cues while maintaining their beneficial interplay, leading to a more human-consistent and robust evaluation framework.

\begin{table*}[htpb]
\centering
\caption{Cross-dataset prediction performance comparison. The best two results in each column are highlighted in boldface.}

\setlength{\tabcolsep}{2mm}
\begin{tabular}{l|ccc|ccc}
\midrule
\multirow{2}{*}{{Methods}} 
& \multicolumn{3}{c|}{{ AIGCIQA2023$^\text{Train} \rightarrow$ AGIQA-3K$^\text{Test}$}} 
& \multicolumn{3}{c}{{AGIQA-3K$^\text{Train} \rightarrow$  AIGCIQA2023$^\text{Test}$}} \\
\cmidrule(lr){2-4} \cmidrule(lr){5-7}
& {Perception} & {Alignment} & { Mean Score} 
& {Perception} & {Alignment} & { Mean Score} \\
\cmidrule{1-7} 
ResNet50 \cite{he2016deep} & 0.576 & 0.473 & 0.525 & 0.599 & 0.432 & 0.516 \\
ViT-B/32 \cite{dosovitskiy2020image} & 0.509 & 0.434 & 0.472 & 0.517 & 0.381 & 0.449 \\
DB-CNN \cite{zhang2018blind} & 0.627 & 0.390 & 0.509 & 0.654 & 0.470 & 0.562 \\
HyperIQA \cite{su2020blindly} & 0.657 & 0.418 & 0.538 & 0.669 & 0.464 & 0.567 \\
TIER \cite{yuan2024tier} & 0.611 & 0.417 & 0.514 & 0.675 & 0.441 & 0.558 \\
IPCE \cite{peng2024aigc} &\textbf{ 0.695} & \textbf{0.591} & \textbf{0.643} & \textbf{0.721} & \textbf{0.553} & \textbf{0.637} \\
\cmidrule{1-7} 
\rowcolor{green!8!white}
ACL-IQA (Ours) & \textbf{0.761} & \textbf{0.676} & \textbf{0.718} & \textbf{0.760} & \textbf{0.632} & \textbf{0.696} \\
\midrule
\end{tabular}
\label{tab:cross-dataset-comparison}
\vspace{-2em}
\end{table*}

\begin{table*}[htbp]
\caption{Cross-dataset performance comparison when transferring from early datasets to the EvalMi-50K. The best two results in each column are highlighted in boldface.}
\centering
\setlength{\tabcolsep}{1.8mm}
\begin{tabular}{l|ccc|ccc}
\toprule
\multirow{2}{*}{\textbf{Method}} & \multicolumn{3}{c|}{\textbf{AGIQA-3K$^{\text{Train}} \rightarrow$ EvalMi-50K$^{\text{Test}}$}} & \multicolumn{3}{c}{\textbf{AIGCIQA2023$^{\text{Train}} \rightarrow$ EvalMi-50K$^{\text{Test}}$}} \\
\cmidrule(lr){2-4} \cmidrule(lr){5-7}
& Perception & Alignment & Mean Score & Perception & Alignment & Mean Score \\
\midrule
ManIQA~\cite{yang2022maniqa} & 0.0619 & -- & 0.0619 & 0.1056 & -- & 0.1056 \\
Re-IQA~\cite{saha2023re} & 0.1791 & -- & 0.1791 & 0.1528 & -- & 0.1528 \\
TReS~\cite{golestaneh2022no} & -0.0677 & -- & -0.0677 & -0.1296 & -- & -0.1296 \\
MA-AGIQA~\cite{wang2024large} & 0.2914 & -- & 0.2914 & 0.4438 & -- & 0.4438 \\
LIQE~\cite{zhang2023blind} & 0.1823 & 0.1857 & 0.1840 & 0.1805 & 0.1568 & 0.1687 \\
IPCE~\cite{peng2024aigc} & \textbf{0.5064} & \textbf{0.3247} & \textbf{0.4156} & \textbf{0.5459} & \textbf{0.4041} & \textbf{0.4750} \\
\midrule
\rowcolor{green!8!white}
\textbf{ACL-IQA (Ours)} & \textbf{0.6057} & \textbf{0.4449} & \textbf{0.5253} & \textbf{0.6567} & \textbf{0.4579} & \textbf{0.5573} \\
\bottomrule
\end{tabular}
\label{tab:cross_dataset_evalmi}
\end{table*}

\begin{table*}[t]
    \caption{Ablation study results on the AGIQA-3K and AIGCIQA2023 datasets. ``Dual LS" denotes the dual (adversarial and collaborative) learning text/instructions. ``Adv." and ``Col." refer to the adversarial and collaborative learning paths, respectively. }
    \centering
    \begin{tabular}{c|ccccc|cc|cc|cc|cc}
        \midrule
        \multirow{4}{*}{{Exp. ID}} & \multirow{4}{*}{{MoE}} & \multirow{4}{*}{{Dual LS}} & \multirow{4}{*}{{Adv.}} & \multirow{4}{*}{{Col.}} & & \multicolumn{4}{c|}{{AGIQA-3K}} & \multicolumn{4}{c}{{AIGCIQA2023}} \\
        \cmidrule(lr){7-10} \cmidrule(lr){11-14}
        & & & & & & \multicolumn{2}{c|}{{Perception}} & \multicolumn{2}{c|}{{Alignment}} & \multicolumn{2}{c|}{{Perception}} & \multicolumn{2}{c}{{Alignment}} \\
        \cmidrule(lr){7-8} \cmidrule(lr){9-10} \cmidrule(lr){11-12} \cmidrule(lr){13-14}
        & & & & & & SRCC & PLCC & SRCC & PLCC & SRCC & PLCC & SRCC & PLCC \\
        \midrule
        1 & \ding{55} & \ding{55} & \ding{55} & \ding{55} & & 0.8841 & 0.9246 & 0.7697 & 0.8725 & 0.8640 & 0.8788 & 0.7979 & 0.7887 \\
        2 & \ding{55} & \ding{55} & $\checkmark$  & $\checkmark$  & & 0.8866 & 0.9253 & 0.7766 & 0.8771 & 0.8677 & 0.8835 & 0.8032 & 0.7939 \\
        3 & $\checkmark$  & \ding{55}   & $\checkmark$  & $\checkmark$  & & 0.8895 & 0.9270 & 0.7787 & 0.8780 & 0.8687 & 0.8839 & 0.8043 & 0.7950 \\
        4 & $\checkmark$ & $\checkmark$  & $\checkmark$  & \ding{55} & & 0.8680 & 0.8682 & 0.7660 & 0.8675 & 0.8622 & 0.8787 & 0.7648 & 0.7545 \\
        5 & $\checkmark$  & $\checkmark$  & \ding{55} & $\checkmark$  & & 0.8606 & 0.9014 & 0.7642 & 0.8673 & 0.8597 & 0.8700 & 0.7701 & 0.7641 \\
         \midrule
         \rowcolor{green!8!white}
        Ours & $\checkmark$  & $\checkmark$  & $\checkmark$  & $\checkmark$  & & \textbf{0.8903} & \textbf{0.9282} & \textbf{0.7832} & \textbf{0.8800} & \textbf{0.8720} & \textbf{0.8865} & \textbf{0.8092} & \textbf{0.8005} \\
        \midrule
    \end{tabular}
    \label{tab:ablation-study}
\end{table*}

\subsubsection{Generalization Ability Comparison}
We further conduct cross-dataset evaluations to assess the generalization capability of ACL-IQA. First, we evaluate the transferability between early-stage datasets such as AGIQA-3K and AIGCIQA2023. As shown in Table~\ref{tab:cross-dataset-comparison}, ACL-IQA achieves the strongest cross-dataset performance among all compared methods. When transferring from AIGCIQA2023 to AGIQA-3K, ACL-IQA obtains SRCCs of 0.761 (perception) and 0.676 (alignment), corresponding to relative improvements of 9.5\% and 14.4\% over the competitive baseline, IPCE. In the reverse direction from AGIQA-3K to AIGCIQA2023, it reaches 0.760 and 0.632, outperforming competitors by 5.4\% and 14.3\%, respectively. These results confirm the robustness and consistency of ACL-IQA across datasets with differing content and distortion characteristics. Notably, even compared with models adopting similar preprocessing pipelines and transformer backbones (e.g., IPCE), ACL-IQA maintains a clear performance margin, indicating that its advantages arise from the proposed learning paradigm rather than architectural or pretraining heuristics.

To further validate whether our framework captures generalizable interaction patterns beyond the priors of early-stage generative models, we conduct a challenging domain-shift evaluation. We train our models on early-stage datasets and evaluate them directly on the EvalMi-50K benchmark. As summarized in Table~\ref{tab:cross_dataset_evalmi}, the significant domain shift from early-stage generators to the modern and diverse generative priors in EvalMi-50K severely impacts the performance of existing methods. However, ACL-IQA demonstrates vastly superior robustness in this setting. In the transfer from AGIQA-3K to EvalMi-50K, our framework significantly elevates perceptual performance compared to MA-AGIQA by raising the SRCC from 0.2914 to 0.6057. Similarly, it consistently outperforms IPCE across both perception and alignment dimensions in these challenging scenarios, indicating that our dual-path mechanism effectively captures intrinsic quality dynamics and successfully scales up to advanced generative systems.

\begin{figure*}[htp]
    \centering
    \setlength{\abovecaptionskip}{2pt}
    \includegraphics[width=1\linewidth]{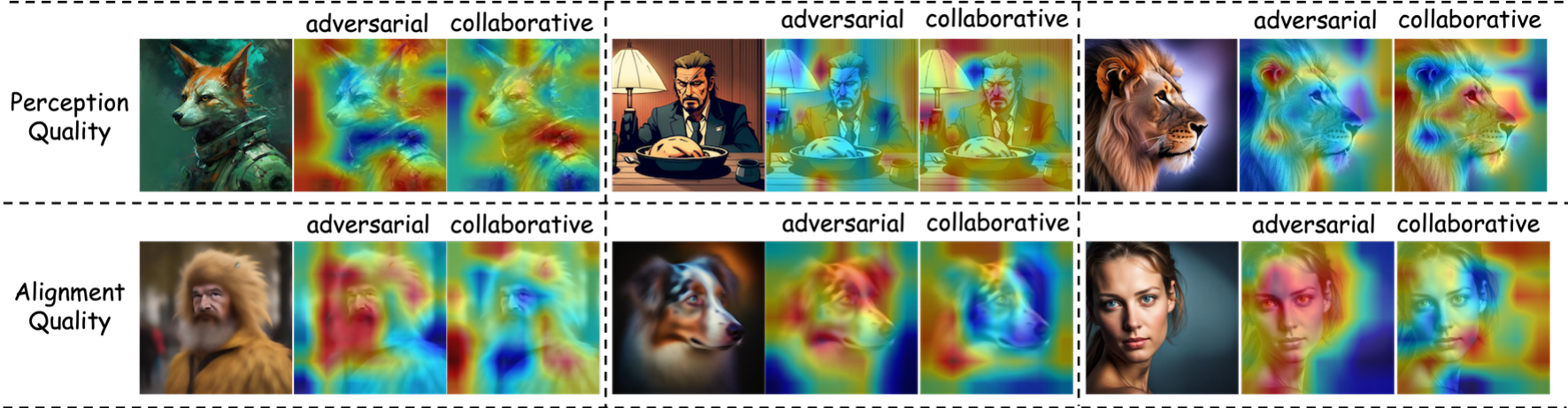}
\caption{Attention visualization of features from the MoE layer. The first and second rows correspond to the perceptual quality and alignment prediction tasks, respectively. 
``Adversarial" and ``Collaborative" indicate that the features are learned from the two distinct interaction paths.
}

    \label{fig:vis_feamap}
\end{figure*}

\begin{figure*}[htbp]
    \includegraphics[width=1.0\linewidth]{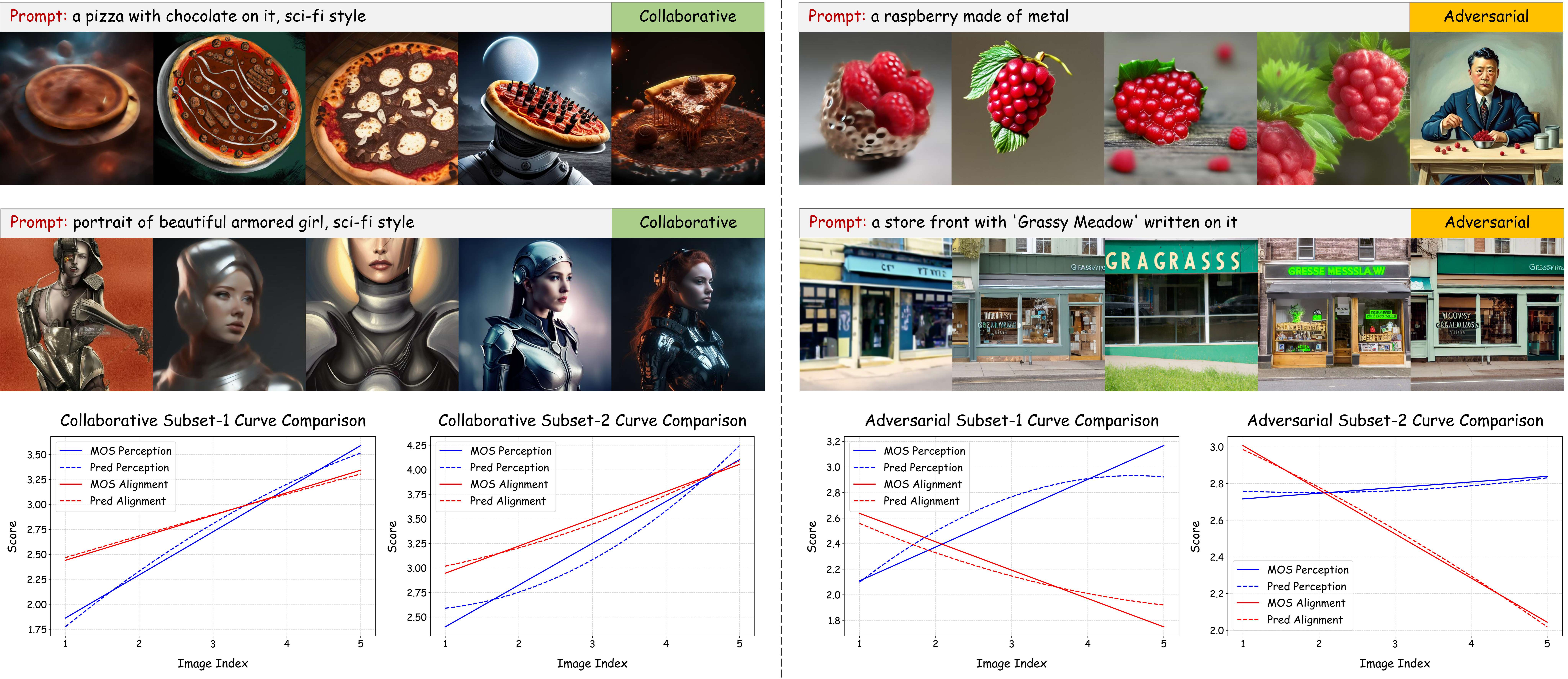}
    \caption{Performance comparison under extreme scenarios, collaborative scenarios (left) where perception quality and alignment quality are positively correlated, and adversarial scenarios (right) where they conflict. Each row displays five generated images alongside their respective prompts. The bottom charts present quantitative comparisons across multiple image sets with perception and alignment quality scores.}
    \label{fig:rank}
\end{figure*}

\begin{figure*}[htbp]
 \centering
    \includegraphics[width=0.91\linewidth]{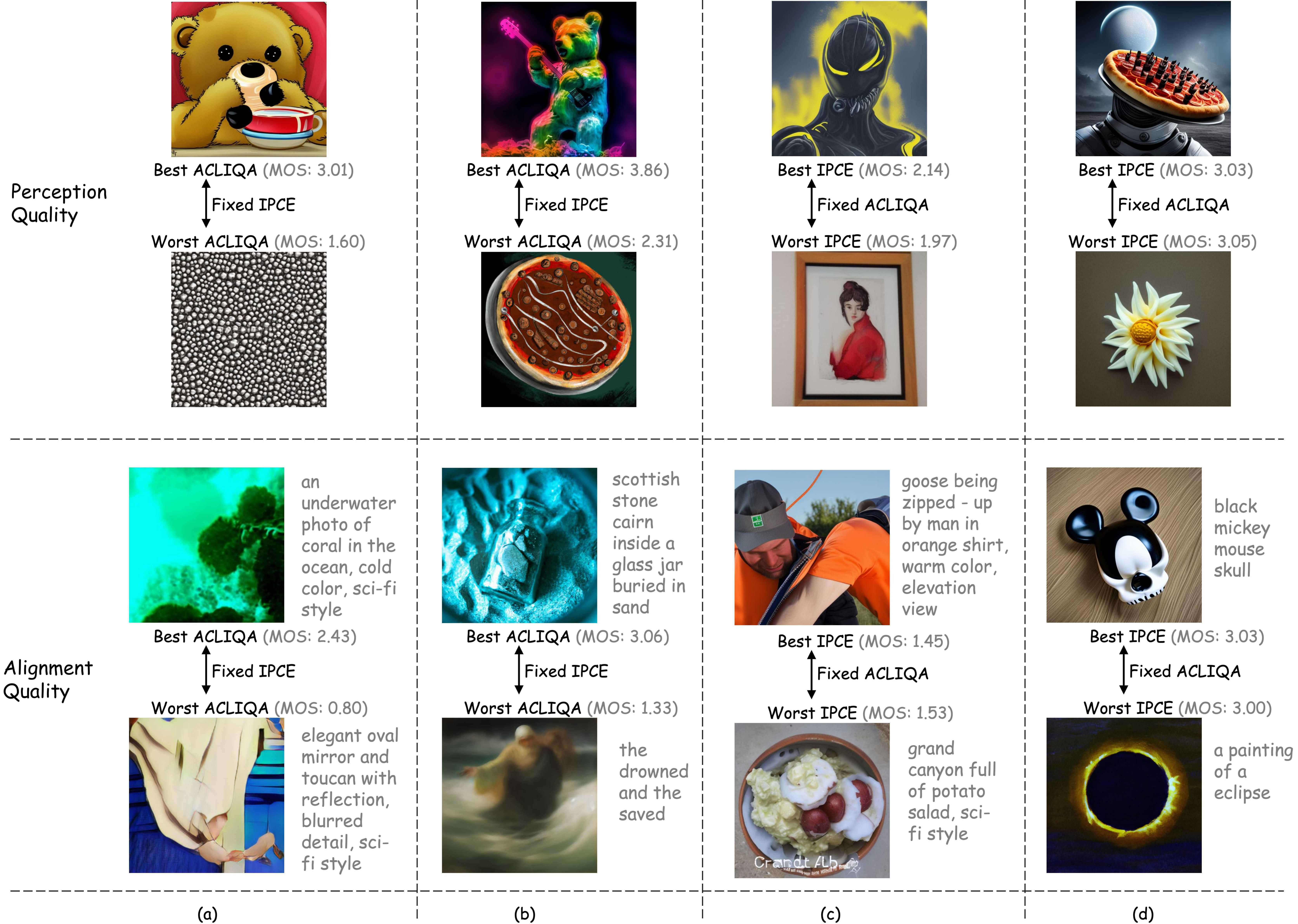}
    \caption{gMAD~\cite{ma2018group} comparison between ACL-IQA and IPCE on perception (top) and alignment (bottom) quality. (a) IPCE predicts both images as low quality, (b) IPCE predicts both images as high quality, (c) both images have low MOS, and (d) both images have high MOS. ACL-IQA aligns more consistently with human judgments across all cases.}
    \label{fig:gmad}
\end{figure*}

\begin{figure}[t]
    \centering
    \setlength{\abovecaptionskip}{2pt}
    \includegraphics[width=1\linewidth]{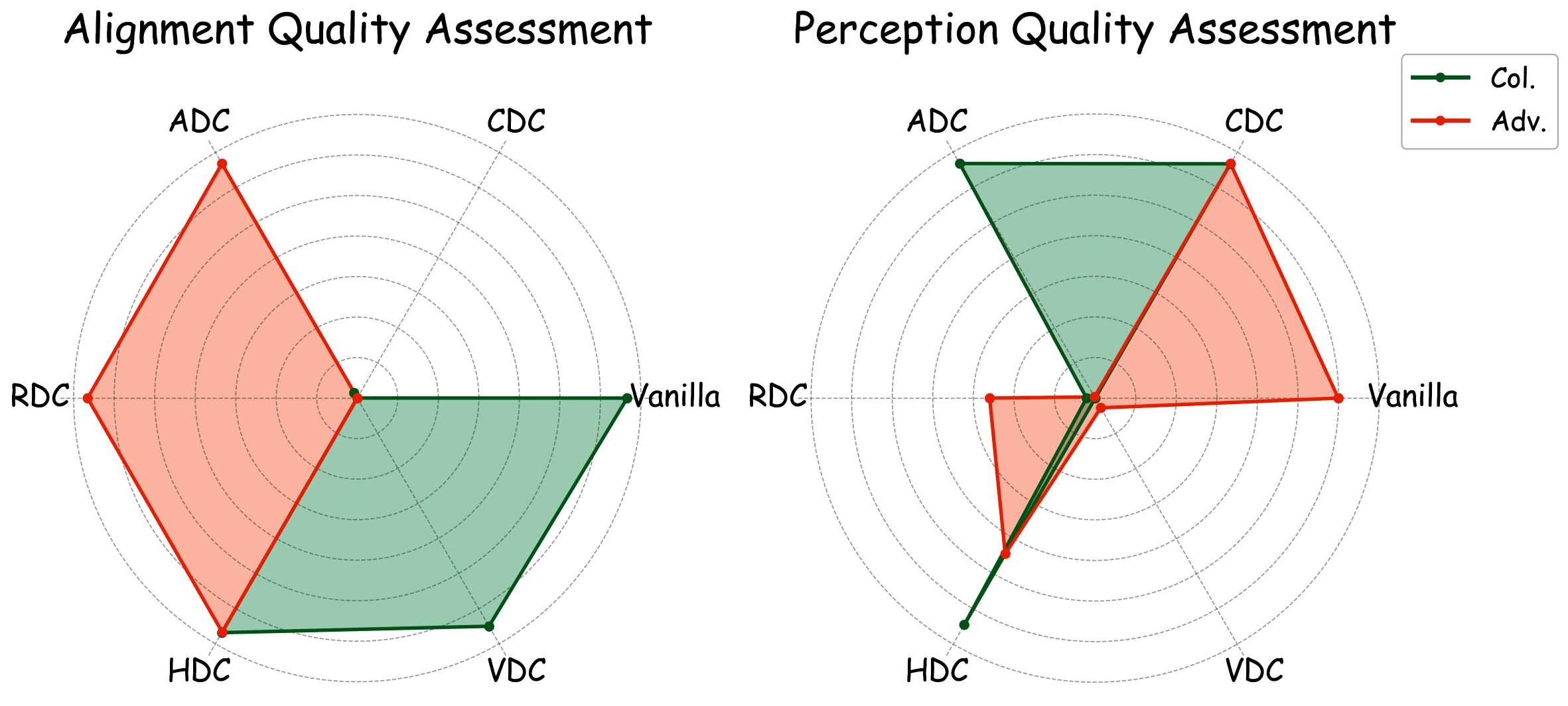}
    \caption{Comparison of gating activations in dual path learning. ``Adv." and ``Col." denote the adversarial and collaborative learning paths, respectively.}
    \label{fig:conv}
\end{figure}

\begin{figure}
    \centering
    \setlength{\abovecaptionskip}{2pt}
    \includegraphics[width=1.0\linewidth]{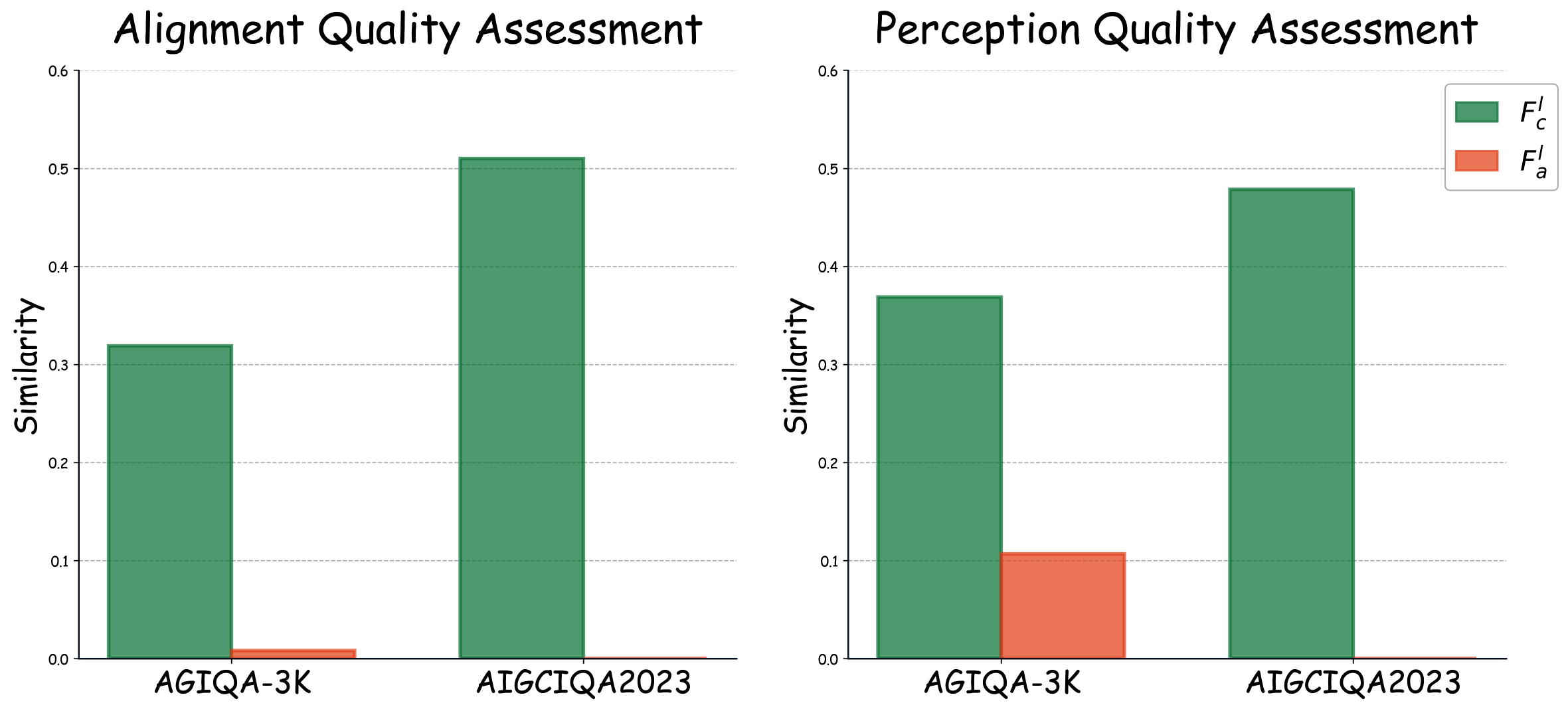}
    \caption{The cosine similarity between the text feature and the image features ($F_{col}^I$, $F_{adv}^I$) extracted in collaborative and adversarial learning.}
    \label{fig:cosim}
\end{figure}

\subsection{Ablation Studies}

The main components proposed in our method include the MoE mechanism, the instruction prompt for dual-path learning, and the Adversarial-Collaborative Mechanism. To assess the effectiveness of each component, we conduct the ablation studies based on five variants of our method, as shown in Table \ref{tab:ablation-study}.  In Exp. 1, we ablate all proposed mechanisms, using only the image encoder and text encoder for quality level classification and quality regression. The results show a significant performance drop, particularly in alignment quality on the AGIQA-3K dataset and perception quality assessment on the AIGCIQA2023 dataset. This highlights the necessity of incorporating the Adversarial-Collaborative Mechanism and MoE for robust quality evaluation. In Exp. 2, we retain the dual-path (Adversarial-Collaborative) learning mechanism but replace the MoE with standard transformer blocks. To maintain dual-path learning, a mapping network is introduced to project image features into two distinct feature representations. The results indicate noticeable improvements over Exp. 1, confirming the effectiveness of the Adversarial-Collaborative Mechanism. However, the performance remains suboptimal compared to the full model, suggesting that MoE also plays a crucial role in refining feature extraction and enhancing the adversarial-collaborative process. In Exp. 3, we examine the contribution of the learning instruction by ablating it from our gating network. A performance drop is observed in the perception assessment on the AGIQA-3K dataset and the alignment assessment on the AIGCIQA2023 dataset, as measured by SRCC. This decline may be attributed to the instruction's role in facilitating fine-grained feature selection for adversarial and cooperative strategies. In Exp. 4 and 5, we evaluate the independent contributions of the adversarial and collaborative mechanisms by applying each separately. The results reveal that while both mechanisms contribute to performance gains, neither outperforms their combination. For instance, the collaborative mechanism shows superior performance in perception quality assessment on AGIQA-3K and alignment quality on AIGCIQA2023, but underperforms in other aspects.  This suggests that adversarial and collaborative strategies capture complementary quality representations. Our full model achieves the best performance across all datasets, confirming that each proposed component benefits our model.

\subsection{Visualization}
To further demonstrate the effectiveness of our model, visualizations of representative feature activations and qualitative results are presented as follows.\\

\noindent \textbf{Path-Specific Attention Feature Maps.} 
As shown in Fig.~\ref{fig:vis_feamap}, our MoE framework captures diverse interaction cues, enhances shared-attention cues, and ablates task-irrelevant attention in collaborative and adversarial learning paths, respectively.
Taking perception quality prediction as an example, the collaborative path attends to regions highly correlated with alignment (\eg, semantically important entities), whereas the adversarial path focuses exclusively on perception-related cues such as sharpness, noise, and artifacts.
Moreover, our MoE layers dynamically select the top-k differential convolutional experts for each path, enabling more targeted feature extraction based on the learning objective.\\

\noindent \textbf{Performance Under Extreme Scenarios.} To demonstrate the robustness of our method, we evaluate its performance under two extreme scenarios: (1) image sets where perception scores are fully correlated with alignment scores, and (2) image sets where perception scores are completely adversarial to alignment scores. The results, shown in Fig.~\ref{fig:rank}, indicate that our method maintains strong consistency with human ratings and outperforms the second-best method, IPCE, in both extreme scenarios. This highlights our method’s superior robustness, as it effectively accounts for both types of interactions observed in human ratings through dual-path learning.\\

\noindent \textbf{Results of gMAD Competition.} To further assess the behavior of different models beyond quantitative metrics, we adopt the generalized Maximum Differentiation (gMAD)~\cite{ma2018group} protocol to perform a head-to-head evaluation against the state-of-the-art IPCE model. As shown in Fig.~\ref{fig:gmad}, both methods are compared on perception and alignment quality across four representative scenarios. In cases where IPCE assigns similarly low or high predictions to paired images, ACL-IQA is still able to correctly differentiate their quality in accordance with MOS, reflecting a higher perceptual sensitivity. Conversely, when the two images share similar MOS but IPCE produces large score discrepancies, ACL-IQA predicts closer values consistent with human judgments. Overall, the results demonstrate that ACL-IQA achieves more reliable and human-aligned behavior in challenging edge cases, highlighting stronger robustness and discriminability compared with IPCE.\\

\noindent \textbf{Gating Activation Comparison in Dual-path Learning.} In our method, the output weights of the experts are determined by our dual-gated network for dual-path learning. To better understand this mechanism,  
we visualize the weight distributions for adversarial and collaborative learning on the AIGCIQA2023 dataset in  Fig.~\ref{fig:conv}. From the figure, we could observe a clear distinction between the expert groups selected by the adversarial and collaborative mechanisms across different assessment tasks. This demonstrates that our designed gating mechanism effectively enables the model to intelligently allocate and learn features based on different learning strategies.\\

\noindent \textbf{Feature Purity Ensured by Adversarial Learning.} In our adversarial learning path, the extracted image features are expected to encode the quality information only in one dimension. To validate the purity of these features and the effectiveness of the adversarial mechanism, we perform inference on the test sets of the AGIQA3K and AIGCIQA2023 datasets, computing the cosine similarity between the text feature and the two image features \(F_{col}^I\) and \(F_{adv}^I\). Taking the alignment quality evaluation task as an example, we first map the ground-truth alignment score to its corresponding quality level \(k\). We then compute the adversarial similarity and collaborative similarity as the cosine similarity between the adversarial feature and the text features representing the \(k\)-th perception quality level and \(k\)-th alignment quality level, respectively.  Finally, the mean similarity values are reported across all test images. Herein, it should be noted that a single quality score usually corresponds to multiple quality levels (\eg, $\hat{Q}_{per}$=3.5 may exhibit high similarity with both the third and fourth quality levels). However, we consider only the most adjacent quality level, which may lead to an underestimation of the similarity score for the collaborative feature \(F_{col}^I\).  As shown in Fig.~\ref{fig:cosim}, the experimental results reveal that the adversarial similarity is significantly lower than the collaborative similarity. This suggests that the adversarial features learned by our model contain minimal quality information from the other dimension, effectively capturing the unique elements that differentiate the two paths.

\section{Conclusion}
In this paper, we re-examined AIGIQA from the perspective of human judgment and demonstrated that perceptual fidelity and prompt alignment do not behave as independent attributes, but exhibit both collaborative and adversarial interactions. To model this property, we proposed a Dual-Gated Mixture-of-Experts framework that dynamically routes features through complementary reasoning paths, enabling adaptive balancing between reinforcement and disentanglement. Extensive experiments across multiple benchmarks show that our approach achieves state-of-the-art performance while delivering more interpretable and human-consistent predictions. We believe this perspective provides new insight that quality assessment models for AI-generated content should move beyond single-factor scoring and instead explicitly account for the interaction patterns underlying human decision making.
\FloatBarrier
\bibliographystyle{IEEEtran}
\bibliography{main}

\begin{thebibliography}{10}
\providecommand{\url}[1]{#1}
\csname url@samestyle\endcsname
\providecommand{\newblock}{\relax}
\providecommand{\bibinfo}[2]{#2}
\providecommand{\BIBentrySTDinterwordspacing}{\spaceskip=0pt\relax}
\providecommand{\BIBentryALTinterwordstretchfactor}{4}
\providecommand{\BIBentryALTinterwordspacing}{\spaceskip=\fontdimen2\font plus
\BIBentryALTinterwordstretchfactor\fontdimen3\font minus \fontdimen4\font\relax}
\providecommand{\BIBforeignlanguage}[2]{{%
\expandafter\ifx\csname l@#1\endcsname\relax
\typeout{** WARNING: IEEEtran.bst: No hyphenation pattern has been}%
\typeout{** loaded for the language `#1'. Using the pattern for}%
\typeout{** the default language instead.}%
\else
\language=\csname l@#1\endcsname
\fi
#2}}
\providecommand{\BIBdecl}{\relax}
\BIBdecl

\bibitem{fang2020perceptual}
Y.~Fang, H.~Zhu, Y.~Zeng, K.~Ma, and Z.~Wang, ``Perceptual quality assessment of smartphone photography,'' in \emph{Proceedings of the IEEE/CVF Conference on Computer Vision and Pattern Recognition}, 2020, pp. 3677--3686.

\bibitem{su2020blindly}
S.~Su, Q.~Yan, Y.~Zhu, C.~Zhang, X.~Ge, J.~Sun, and Y.~Zhang, ``Blindly assess image quality in the wild guided by a self-adaptive hyper network,'' in \emph{Proceedings of the IEEE/CVF Conference on Computer Vision and Pattern Recognition}, 2020, pp. 3667--3676.

\bibitem{min2017unified}
X.~Min, K.~Ma, K.~Gu, G.~Zhai, Z.~Wang, and W.~Lin, ``Unified blind quality assessment of compressed natural, graphic, and screen content images,'' \emph{IEEE Transactions on Image Processing}, vol.~26, no.~11, pp. 5462--5474, 2017.

\bibitem{chen2021learning}
B.~Chen, L.~Zhu, G.~Li, F.~Lu, H.~Fan, and S.~Wang, ``{Learning generalized spatial-temporal deep feature representation for no-reference video quality assessment},'' \emph{IEEE Transactions on Circuits and Systems for Video Technology}, vol.~32, no.~4, pp. 1903--1916, 2021.

\bibitem{chen2025monotonic}
B.~Chen, K.~Xiao, X.~Shen, and S.~Wang, ``Monotonic and invertible network: A general framework for learning iqa model from mixed datasets,'' \emph{International Journal of Computer Vision}, vol. 133, no.~11, pp. 7924--7945, 2025.

\bibitem{chen2026global}
B.~Chen, D.~Huang, H.~Zhu, L.~Zhu, W.~Zhou, S.~Wang, Y.~Fang, and W.~Lin, ``{From global to granular: Revealing iqa model performance via correlation surface},'' \emph{IEEE Transactions on Pattern Analysis and Machine Intelligence}, 2026.

\bibitem{yuan2024tier}
J.~Yuan, X.~Cao, J.~Che, Q.~Wang, S.~Liang, W.~Ren, J.~Lin, and X.~Cao, ``{TIER}: Text-image encoder-based regression for aigc image quality assessment,'' \emph{arXiv preprint arXiv:2401.03854}, 2024.

\bibitem{peng2024aigc}
F.~Peng, H.~Fu, A.~Ming, C.~Wang, H.~Ma, S.~He, Z.~Dou, and S.~Chen, ``{AIGC} image quality assessment via image-prompt correspondence,'' in \emph{Proceedings of the IEEE/CVF Conference on Computer Vision and Pattern Recognition}, 2024, pp. 6432--6441.

\bibitem{radford2021learning}
A.~Radford, J.~W. Kim, C.~Hallacy, A.~Ramesh, G.~Goh, S.~Agarwal, G.~Sastry, A.~Askell, P.~Mishkin, J.~Clark \emph{et~al.}, ``Learning transferable visual models from natural language supervision,'' in \emph{International Conference on Machine Learning}.\hskip 1em plus 0.5em minus 0.4em\relax PmLR, 2021, pp. 8748--8763.

\bibitem{goodfellow2020generative}
I.~Goodfellow, J.~Pouget-Abadie, M.~Mirza, B.~Xu, D.~Warde-Farley, S.~Ozair, A.~Courville, and Y.~Bengio, ``Generative adversarial networks,'' \emph{Communications of the ACM}, vol.~63, no.~11, pp. 139--144, 2020.

\bibitem{ding2021cogview}
M.~Ding, Z.~Yang, W.~Hong, W.~Zheng, C.~Zhou, D.~Yin, J.~Lin, X.~Zou, Z.~Shao, H.~Yang \emph{et~al.}, ``{CogView}: Mastering text-to-image generation via transformers,'' \emph{Advances in Neural Information Processing Systems}, vol.~34, pp. 19\,822--19\,835, 2021.

\bibitem{rombach2022high}
R.~Rombach, A.~Blattmann, D.~Lorenz, P.~Esser, and B.~Ommer, ``High-resolution image synthesis with latent diffusion models,'' in \emph{Proceedings of the IEEE/CVF Conference on Computer Vision and Pattern Recognition}, 2022, pp. 10\,684--10\,695.

\bibitem{zhang2023perceptual}
Z.~Zhang, C.~Li, W.~Sun, X.~Liu, X.~Min, and G.~Zhai, ``A perceptual quality assessment exploration for aigc images,'' in \emph{2023 IEEE International Conference on Multimedia and Expo Workshops (ICMEW)}.\hskip 1em plus 0.5em minus 0.4em\relax IEEE, 2023, pp. 440--445.

\bibitem{li2023agiqa}
C.~Li, Z.~Zhang, H.~Wu, W.~Sun, X.~Min, X.~Liu, G.~Zhai, and W.~Lin, ``{AGIQA-3k}: An open database for ai-generated image quality assessment,'' \emph{IEEE Transactions on Circuits and Systems for Video Technology}, vol.~34, no.~8, pp. 6833--6846, 2023.

\bibitem{zhu2025adaptive}
H.~Zhu, H.~Wu, Y.~Li, Z.~Zhang, B.~Chen, L.~Zhu, Y.~Fang, G.~Zhai, W.~Lin, and S.~Wang, ``Adaptive image quality assessment via teaching large multimodal model to compare,'' \emph{Advances in Neural Information Processing Systems}, vol.~37, pp. 32\,611--32\,629, 2025.

\bibitem{wang2023aigciqa2023}
J.~Wang, H.~Duan, J.~Liu, S.~Chen, X.~Min, and G.~Zhai, ``{AIGCIQA2023}: A large-scale image quality assessment database for ai generated images: from the perspectives of quality, authenticity and correspondence,'' in \emph{CAAI International Conference on Artificial Intelligence}.\hskip 1em plus 0.5em minus 0.4em\relax Springer, 2023, pp. 46--57.

\bibitem{li2024aigiqa}
C.~Li, T.~Kou, Y.~Gao, Y.~Cao, W.~Sun, Z.~Zhang, Y.~Zhou, Z.~Zhang, W.~Zhang, H.~Wu \emph{et~al.}, ``{AIGIQA-20K}: A large database for ai-generated image quality assessment,'' in \emph{Proceedings of the IEEE/CVF Conference on Computer Vision and Pattern Recognition}, 2024, pp. 6327--6336.

\bibitem{wang2025lmm4lmm}
J.~Wang, H.~Duan, Y.~Zhao, J.~Wang, G.~Zhai, and X.~Min, ``{LMM4LMM}: Benchmarking and evaluating large-multimodal image generation with lmms,'' in \emph{Proceedings of the IEEE/CVF International Conference on Computer Vision}, 2025, pp. 17\,312--17\,323.

\bibitem{han2024evalmuse}
S.~Han, H.~Fan, J.~Fu, L.~Li, T.~Li, J.~Cui, Y.~Wang, Y.~Tai, J.~Sun, C.~Guo \emph{et~al.}, ``{EvalMuse-40K}: A reliable and fine-grained benchmark with comprehensive human annotations for text-to-image generation model evaluation,'' \emph{arXiv preprint arXiv:2412.18150}, 2024.

\bibitem{li2024genai}
B.~Li, Z.~Lin, D.~Pathak, J.~Li, Y.~Fei, K.~Wu, T.~Ling, X.~Xia, P.~Zhang, G.~Neubig \emph{et~al.}, ``{GenAI-Bench}: Evaluating and improving compositional text-to-visual generation,'' \emph{arXiv preprint arXiv:2406.13743}, 2024.

\bibitem{huang2023t2i}
K.~Huang, K.~Sun, E.~Xie, Z.~Li, and X.~Liu, ``{T2I-CompBench}: A comprehensive benchmark for open-world compositional text-to-image generation,'' \emph{Advances in Neural Information Processing Systems}, vol.~36, pp. 78\,723--78\,747, 2023.

\bibitem{ghosh2023geneval}
D.~Ghosh, H.~Hajishirzi, and L.~Schmidt, ``{GenEval}: An object-focused framework for evaluating text-to-image alignment,'' \emph{Advances in Neural Information Processing Systems}, vol.~36, pp. 52\,132--52\,152, 2023.

\bibitem{liang2024rich}
Y.~Liang, J.~He, G.~Li, P.~Li, A.~Klimovskiy, N.~Carolan, J.~Sun, J.~Pont-Tuset, S.~Young, F.~Yang \emph{et~al.}, ``Rich human feedback for text-to-image generation,'' in \emph{CVPR}, 2024.

\bibitem{salimans2016improved}
T.~Salimans, I.~Goodfellow, W.~Zaremba, V.~Cheung, A.~Radford, and X.~Chen, ``Improved techniques for training gans,'' \emph{Advances in Neural Information Processing Systems}, vol.~29, 2016.

\bibitem{heusel2017gans}
M.~Heusel, H.~Ramsauer, T.~Unterthiner, B.~Nessler, and S.~Hochreiter, ``Gans trained by a two time-scale update rule converge to a local nash equilibrium,'' in \emph{Advances in Neural Information Processing Systems}, vol.~30, 2017.

\bibitem{zhu2020metaiqa}
H.~Zhu, L.~Li, J.~Wu, W.~Dong, and G.~Shi, ``{MetaIQA}: Deep meta-learning for no-reference image quality assessment,'' in \emph{Proceedings of the IEEE/CVF conference on computer vision and pattern recognition}, 2020, pp. 14\,143--14\,152.

\bibitem{sun2022graphiqa}
S.~Sun, T.~Yu, J.~Xu, W.~Zhou, and Z.~Chen, ``{GraphIQA}: Learning distortion graph representations for blind image quality assessment,'' \emph{IEEE Transactions on Multimedia}, vol.~25, pp. 2912--2925, 2022.

\bibitem{fang2024pcqa}
X.~Fang, W.~Wang, X.~Lv, and J.~Yan, ``{PCQA}: A strong baseline for {AIGC} quality assessment based on prompt condition,'' in \emph{Proceedings of the IEEE/CVF Conference on Computer Vision and Pattern Recognition}, 2024, pp. 6167--6176.

\bibitem{fu2024vision}
J.~Fu, W.~Zhou, Q.~Jiang, H.~Liu, and G.~Zhai, ``Vision-language consistency guided multi-modal prompt learning for blind {AI} generated image quality assessment,'' \emph{IEEE Signal Processing Letters}, 2024.

\bibitem{golestaneh2022no}
S.~A. Golestaneh, S.~Dadsetan, and K.~M. Kitani, ``No-reference image quality assessment via transformers, relative ranking, and self-consistency,'' in \emph{Proceedings of the IEEE/CVF winter Conference on Applications of Computer Vision}, 2022, pp. 1220--1230.

\bibitem{yang2022maniqa}
S.~Yang, T.~Wu, S.~Shi, S.~Lao, Y.~Gong, M.~Cao, J.~Wang, and Y.~Yang, ``{MANIQA}: Multi-dimension attention network for no-reference image quality assessment,'' in \emph{Proceedings of the IEEE/CVF conference on computer vision and pattern recognition}, 2022, pp. 1191--1200.

\bibitem{saha2023re}
A.~Saha, S.~Mishra, and A.~C. Bovik, ``{Re-IQA}: Unsupervised learning for image quality assessment in the wild,'' in \emph{Proceedings of the IEEE/CVF Conference on Computer Vision and Pattern Recognition}, 2023, pp. 5846--5855.

\bibitem{zhang2023blind}
W.~Zhang, G.~Zhai, Y.~Wei, X.~Yang, and K.~Ma, ``Blind image quality assessment via vision-language correspondence: A multitask learning perspective,'' in \emph{Proceedings of the IEEE/CVF Conference on Computer Vision and Pattern Recognition}, 2023, pp. 14\,071--14\,081.

\bibitem{qu2024bringing}
B.~Qu, H.~Li, and W.~Gao, ``Bringing textual prompt to {AI}-generated image quality assessment,'' in \emph{IEEE International Conference on Multimedia and Expo}, 2024, pp. 1--6.

\bibitem{yu2024sf}
Z.~Yu, F.~Guan, Y.~Lu, X.~Li, and Z.~Chen, ``{SF-IQA}: Quality and similarity integration for ai generated image quality assessment,'' in \emph{Proceedings of the IEEE/CVF Conference on Computer Vision and Pattern Recognition}, 2024, pp. 6692--6701.

\bibitem{xia2024ai}
J.~Xia, L.~He, F.~Gao, K.~Zhang, L.~Li, and X.~Gao, ``{AI}-generated image quality assessment based on task-specific prompt and multi-granularity similarity,'' \emph{arXiv preprint arXiv:2411.16087}, 2024.

\bibitem{meng2026decoupling}
Z.~Meng, ``{Decoupling Semantics from Distortions: Multi-Scale Two-Stream Vision-Language Alignment for AI-Generated Image Quality Assessment},'' \emph{arXiv preprint arXiv:2606.16799}, 2026.

\bibitem{zhao2025reasoning}
S.~Zhao, X.~Zhang, W.~Li, J.~Li, L.~Zhang, T.~Xue, and J.~Zhang, ``Reasoning as representation: Rethinking visual reinforcement learning in image quality assessment,'' \emph{arXiv preprint arXiv:2510.11369}, 2025.

\bibitem{liao2025plug}
Z.~Liao, B.~Chen, H.~Zhu, L.~Zhu, S.~Wang, and W.~Lin, ``{Plug In, Grade Right: Psychology-Inspired AGIQA},'' \emph{arXiv preprint arXiv:2512.22780}, 2025.

\bibitem{li2025text}
Q.~Li, Q.~Yan, H.~Huang, P.~Wu, H.~Zhang, and Y.~Zhang, ``{Text-visual semantic constrained AI-generated image quality assessment},'' in \emph{Proceedings of the 33rd ACM International Conference on Multimedia}, 2025, pp. 6958--6966.

\bibitem{wu2024qalign}
H.~Wu, Z.~Zhang, W.~Zhang, C.~Chen, L.~Liao, C.~Li, Y.~Gao, A.~Wang, E.~Zhang, W.~Sun \emph{et~al.}, ``{Q-Align}: Teaching lmms for visual scoring via discrete text-defined levels,'' in \emph{Proceedings of the 41st International Conference on Machine Learning (ICML)}, vol. 235, 2024, pp. 54\,015--54\,029.

\bibitem{wang2024large}
P.~Wang, W.~Sun, Z.~Zhang, J.~Jia, Y.~Jiang, Z.~Zhang, X.~Min, and G.~Zhai, ``Large multi-modality model assisted ai-generated image quality assessment,'' in \emph{Proceedings of the 32nd ACM International Conference on Multimedia}, 2024, pp. 7803--7812.

\bibitem{you2025teaching}
Z.~You, X.~Cai, J.~Gu, T.~Xue, and C.~Dong, ``Teaching large language models to regress accurate image quality scores using score distribution,'' in \emph{Proceedings of the Computer Vision and Pattern Recognition Conference}, 2025, pp. 14\,483--14\,494.

\bibitem{li2026q}
W.~Li, X.~Zhang, S.~Zhao, Y.~Zhang, J.~Li, J.~Zhang \emph{et~al.}, ``{Q-Insight}: Understanding image quality via visual reinforcement learning,'' \emph{Advances in Neural Information Processing Systems}, vol.~38, pp. 36\,802--36\,827, 2026.

\bibitem{jacobs1991adaptive}
R.~A. Jacobs, M.~I. Jordan, S.~J. Nowlan, and G.~E. Hinton, ``Adaptive mixtures of local experts,'' \emph{Neural computation}, vol.~3, no.~1, pp. 79--87, 1991.

\bibitem{shazeer2017outrageously}
N.~Shazeer, A.~Mirhoseini, K.~Maziarz, A.~Davis, Q.~Le, G.~Hinton, and J.~Dean, ``Outrageously large neural networks: The sparsely-gated mixture-of-experts layer,'' \emph{arXiv preprint arXiv:1701.06538}, 2017.

\bibitem{fedus2022switch}
W.~Fedus, B.~Zoph, and N.~Shazeer, ``Switch transformers: Scaling to trillion parameter models with simple and efficient sparsity,'' \emph{Journal of Machine Learning Research}, vol.~23, no. 120, pp. 1--39, 2022.

\bibitem{kudugunta2021beyond}
S.~Kudugunta, Y.~Huang, A.~Bapna, M.~Krikun, D.~Lepikhin, M.-T. Luong, and O.~Firat, ``Beyond distillation: Task-level mixture-of-experts for efficient inference,'' \emph{arXiv preprint arXiv:2110.03742}, 2021.

\bibitem{ma2018modeling}
J.~Ma, Z.~Zhao, X.~Yi, J.~Chen, L.~Hong, and E.~H. Chi, ``Modeling task relationships in multi-task learning with multi-gate mixture-of-experts,'' in \emph{Proceedings of the ACM SIGKDD International Conference on Knowledge Discovery \& Data mining}, 2018, pp. 1930--1939.

\bibitem{aoki2022heterogeneous}
R.~Aoki, F.~Tung, and G.~L. Oliveira, ``Heterogeneous multi-task learning with expert diversity,'' \emph{IEEE/ACM Transactions on Computational Biology and Bioinformatics}, vol.~19, no.~6, pp. 3093--3102, 2022.

\bibitem{fan2022m3vit}
Z.~Fan, R.~Sarkar, Z.~Jiang, T.~Chen, K.~Zou, Y.~Cheng, C.~Hao, Z.~Wang \emph{et~al.}, ``{M$^3$Vit}: Mixture-of-experts vision transformer for efficient multi-task learning with model-accelerator co-design,'' \emph{Advances in Neural Information Processing Systems}, vol.~35, pp. 28\,441--28\,457, 2022.

\bibitem{dosovitskiy2020image}
\BIBentryALTinterwordspacing
A.~Dosovitskiy, L.~Beyer, A.~Kolesnikov, D.~Weissenborn, X.~Zhai, T.~Unterthiner, M.~Dehghani, M.~Minderer, G.~Heigold, S.~Gelly, J.~Uszkoreit, and N.~Houlsby, ``An image is worth 16x16 words: Transformers for image recognition at scale,'' in \emph{International Conference on Learning Representations}, 2021. [Online]. Available: \url{https://openreview.net/forum?id=YicbFdNTTy}
\BIBentrySTDinterwordspacing

\bibitem{su2021pixel}
Z.~Su, W.~Liu, Z.~Yu, D.~Hu, Q.~Liao, Q.~Tian, M.~Pietik{\"a}inen, and L.~Liu, ``Pixel difference networks for efficient edge detection,'' in \emph{Proceedings of the IEEE/CVF International Conference on Computer Vision}, 2021, pp. 5117--5127.

\bibitem{kong2024moe}
C.~Kong, A.~Luo, P.~Bao, Y.~Yu, H.~Li, Z.~Zheng, S.~Wang, and A.~C. Kot, ``{MoE-FFD}: Mixture of experts for generalized and parameter-efficient face forgery detection,'' \emph{arXiv preprint arXiv:2404.08452}, 2024.

\bibitem{park2022how}
N.~Park and S.~Kim, ``How do vision transformers work?'' in \emph{International Conference on Learning Representations}, 2022.

\bibitem{ganin2015unsupervised}
Y.~Ganin and V.~Lempitsky, ``Unsupervised domain adaptation by backpropagation,'' in \emph{International Conference on Machine Learning}.\hskip 1em plus 0.5em minus 0.4em\relax PMLR, 2015, pp. 1180--1189.

\bibitem{kang2014convolutional}
L.~Kang, P.~Ye, Y.~Li, and D.~Doermann, ``Convolutional neural networks for no-reference image quality assessment,'' in \emph{Proceedings of the IEEE Conference on Computer Vision and Pattern Recognition}, 2014, pp. 1733--1740.

\bibitem{zhang2018blind}
W.~Zhang, K.~Ma, J.~Yan, D.~Deng, and Z.~Wang, ``Blind image quality assessment using a deep bilinear convolutional neural network,'' \emph{IEEE Transactions on Circuits and Systems for Video Technology}, vol.~30, no.~1, pp. 36--47, 2018.

\bibitem{wang2021multi}
T.~Wang, W.~Sun, X.~Min, W.~Lu, Z.~Zhang, and G.~Zhai, ``A multi-dimensional aesthetic quality assessment model for mobile game images,'' in \emph{International Conference on Visual Communications and Image Processing (VCIP)}, 2021, pp. 1--5.

\bibitem{wang2023exploring}
J.~Wang, K.~C. Chan, and C.~C. Loy, ``Exploring clip for assessing the look and feel of images,'' in \emph{Proceedings of the AAAI Conference on Artificial Intelligence}, vol.~37, no.~2, 2023, pp. 2555--2563.

\bibitem{sun2023blind}
W.~Sun, X.~Min, D.~Tu, S.~Ma, and G.~Zhai, ``Blind quality assessment for in-the-wild images via hierarchical feature fusion and iterative mixed database training,'' \emph{IEEE Journal of Selected Topics in Signal Processing}, 2023.

\bibitem{yuan2023pscr}
J.~Yuan, X.~Cao, L.~Cao, J.~Lin, and X.~Cao, ``{PSCR}: Patches sampling-based contrastive regression for aigc image quality assessment,'' \emph{arXiv preprint arXiv:2312.05897}, 2023.

\bibitem{xu2023imagereward}
J.~Xu, X.~Liu, Y.~Wu, Y.~Tong, Q.~Li, M.~Ding, J.~Tang, and Y.~Dong, ``{ImageReward}: Learning and evaluating human preferences for text-to-image generation,'' \emph{Advances in Neural Information Processing Systems}, vol.~36, pp. 15\,903--15\,935, 2023.

\bibitem{wu2023human}
X.~Wu, K.~Sun, F.~Zhu, R.~Zhao, and H.~Li, ``Human preference score: Better aligning text-to-image models with human preference,'' in \emph{Proceedings of the IEEE/CVF International Conference on Computer Vision}, 2023, pp. 2096--2105.

\bibitem{kirstain2023pick}
Y.~Kirstain, A.~Polyak, U.~Singer, S.~Matiana, J.~Penna, and O.~Levy, ``{Pick-a-Pic}: An open dataset of user preferences for text-to-image generation,'' \emph{Advances in Neural Information Processing Systems}, vol.~36, pp. 36\,652--36\,663, 2023.

\bibitem{tang2024clip}
Z.~Tang, Z.~Wang, B.~Peng, and J.~Dong, ``{CLIP-AGIQA}: boosting the performance of ai-generated image quality assessment with clip,'' in \emph{International Conference on Pattern Recognition}.\hskip 1em plus 0.5em minus 0.4em\relax Springer, 2024, pp. 48--61.

\bibitem{he2016deep}
K.~He, X.~Zhang, S.~Ren, and J.~Sun, ``Deep residual learning for image recognition,'' in \emph{Proceedings of the IEEE Conference on Computer Vision and Pattern Recognition}, 2016, pp. 770--778.

\bibitem{ke2021musiq}
J.~Ke, Q.~Wang, Y.~Wang, P.~Milanfar, and F.~Yang, ``{MUSIQ}: Multi-scale image quality transformer,'' in \emph{Proceedings of the IEEE/CVF International Conference on Computer Vision}, 2021, pp. 5148--5157.

\bibitem{wu2024deepseek}
Z.~Wu, X.~Chen, Z.~Pan, X.~Liu, W.~Liu, D.~Dai, H.~Gao, Y.~Ma, C.~Wu, B.~Wang \emph{et~al.}, ``{DeepSeek-VL2}: Mixture-of-experts vision-language models for advanced multimodal understanding,'' \emph{arXiv preprint arXiv:2412.10302}, 2024.

\bibitem{wang2024qwen2}
P.~Wang, S.~Bai, S.~Tan, S.~Wang, Z.~Fan, J.~Bai, K.~Chen, X.~Liu, J.~Wang, W.~Ge \emph{et~al.}, ``{Qwen2-VL}: Enhancing vision-language model's perception of the world at any resolution,'' \emph{arXiv preprint arXiv:2409.12191}, 2024.

\bibitem{meta2024llama}
A.~Meta, ``{Llama 3.2}: Revolutionizing edge ai and vision with open, customizable models,'' \emph{Meta AI Blog. Retrieved December}, vol.~20, p. 2024, 2024.

\bibitem{ma2018group}
K.~Ma, Z.~Duanmu, Z.~Wang, Q.~Wu, W.~Liu, H.~Yong, H.~Li, and L.~Zhang, ``Group maximum differentiation competition: Model comparison with few samples,'' \emph{IEEE Transactions on Pattern Analysis and Machine Intelligence}, vol.~42, no.~4, pp. 851--864, 2018.

\end{thebibliography}


\begin{thebibliography}{10}
\providecommand{\url}[1]{#1}
\csname url@samestyle\endcsname
\providecommand{\newblock}{\relax}
\providecommand{\bibinfo}[2]{#2}
\providecommand{\BIBentrySTDinterwordspacing}{\spaceskip=0pt\relax}
\providecommand{\BIBentryALTinterwordstretchfactor}{4}
\providecommand{\BIBentryALTinterwordspacing}{\spaceskip=\fontdimen2\font plus
\BIBentryALTinterwordstretchfactor\fontdimen3\font minus \fontdimen4\font\relax}
\providecommand{\BIBforeignlanguage}[2]{{%
\expandafter\ifx\csname l@#1\endcsname\relax
\typeout{** WARNING: IEEEtran.bst: No hyphenation pattern has been}%
\typeout{** loaded for the language `#1'. Using the pattern for}%
\typeout{** the default language instead.}%
\else
\language=\csname l@#1\endcsname
\fi
#2}}
\providecommand{\BIBdecl}{\relax}
\BIBdecl

\bibitem{han2024evalmuse}
S.~Han, H.~Fan, J.~Fu, L.~Li, T.~Li, J.~Cui, Y.~Wang, Y.~Tai, J.~Sun, C.~Guo \emph{et~al.}, ``{EvalMuse-40K}: A reliable and fine-grained benchmark with comprehensive human annotations for text-to-image generation model evaluation,'' \emph{arXiv preprint arXiv:2412.18150}, 2024.

\bibitem{li2024genai}
B.~Li, Z.~Lin, D.~Pathak, J.~Li, Y.~Fei, K.~Wu, T.~Ling, X.~Xia, P.~Zhang, G.~Neubig \emph{et~al.}, ``{GenAI-Bench}: Evaluating and improving compositional text-to-visual generation,'' \emph{arXiv preprint arXiv:2406.13743}, 2024.

\bibitem{liang2024rich}
Y.~Liang, J.~He, G.~Li, P.~Li, A.~Klimovskiy, N.~Carolan, J.~Sun, J.~Pont-Tuset, S.~Young, F.~Yang \emph{et~al.}, ``Rich human feedback for text-to-image generation,'' in \emph{CVPR}, 2024.

\bibitem{su2021pixel}
Z.~Su, W.~Liu, Z.~Yu, D.~Hu, Q.~Liao, Q.~Tian, M.~Pietik{\"a}inen, and L.~Liu, ``Pixel difference networks for efficient edge detection,'' in \emph{Proceedings of the IEEE/CVF International Conference on Computer Vision}, 2021, pp. 5117--5127.

\bibitem{kong2024moe}
C.~Kong, A.~Luo, P.~Bao, Y.~Yu, H.~Li, Z.~Zheng, S.~Wang, and A.~C. Kot, ``{MoE-FFD}: Mixture of experts for generalized and parameter-efficient face forgery detection,'' \emph{arXiv preprint arXiv:2404.08452}, 2024.

\bibitem{zhang2023blind}
W.~Zhang, G.~Zhai, Y.~Wei, X.~Yang, and K.~Ma, ``Blind image quality assessment via vision-language correspondence: A multitask learning perspective,'' in \emph{Proceedings of the IEEE/CVF Conference on Computer Vision and Pattern Recognition}, 2023, pp. 14\,071--14\,081.

\bibitem{wu2024qalign}
H.~Wu, Z.~Zhang, W.~Zhang, C.~Chen, L.~Liao, C.~Li, Y.~Gao, A.~Wang, E.~Zhang, W.~Sun \emph{et~al.}, ``{Q-Align}: Teaching lmms for visual scoring via discrete text-defined levels,'' in \emph{Proceedings of the 41st International Conference on Machine Learning (ICML)}, vol. 235, 2024, pp. 54\,015--54\,029.

\bibitem{wang2025lmm4lmm}
J.~Wang, H.~Duan, Y.~Zhao, J.~Wang, G.~Zhai, and X.~Min, ``{LMM4LMM}: Benchmarking and evaluating large-multimodal image generation with lmms,'' in \emph{Proceedings of the IEEE/CVF International Conference on Computer Vision}, 2025, pp. 17\,312--17\,323.

\bibitem{li2023agiqa}
C.~Li, Z.~Zhang, H.~Wu, W.~Sun, X.~Min, X.~Liu, G.~Zhai, and W.~Lin, ``{AGIQA-3k}: An open database for ai-generated image quality assessment,'' \emph{IEEE Transactions on Circuits and Systems for Video Technology}, vol.~34, no.~8, pp. 6833--6846, 2023.

\bibitem{wang2023aigciqa2023}
J.~Wang, H.~Duan, J.~Liu, S.~Chen, X.~Min, and G.~Zhai, ``{AIGCIQA2023}: A large-scale image quality assessment database for ai generated images: from the perspectives of quality, authenticity and correspondence,'' in \emph{CAAI International Conference on Artificial Intelligence}.\hskip 1em plus 0.5em minus 0.4em\relax Springer, 2023, pp. 46--57.

\bibitem{zhang2025qwen3}
Y.~Zhang, M.~Li, D.~Long, X.~Zhang, H.~Lin, B.~Yang, P.~Xie, A.~Yang, D.~Liu, J.~Lin \emph{et~al.}, ``{Qwen3 Embedding}: Advancing text embedding and reranking through foundation models,'' \emph{arXiv preprint arXiv:2506.05176}, 2025.

\bibitem{radford2021learning}
A.~Radford, J.~W. Kim, C.~Hallacy, A.~Ramesh, G.~Goh, S.~Agarwal, G.~Sastry, A.~Askell, P.~Mishkin, J.~Clark \emph{et~al.}, ``Learning transferable visual models from natural language supervision,'' in \emph{International Conference on Machine Learning}.\hskip 1em plus 0.5em minus 0.4em\relax PmLR, 2021, pp. 8748--8763.

\bibitem{dosovitskiy2020image}
\BIBentryALTinterwordspacing
A.~Dosovitskiy, L.~Beyer, A.~Kolesnikov, D.~Weissenborn, X.~Zhai, T.~Unterthiner, M.~Dehghani, M.~Minderer, G.~Heigold, S.~Gelly, J.~Uszkoreit, and N.~Houlsby, ``An image is worth 16x16 words: Transformers for image recognition at scale,'' in \emph{International Conference on Learning Representations}, 2021. [Online]. Available: \url{https://openreview.net/forum?id=YicbFdNTTy}
\BIBentrySTDinterwordspacing

\bibitem{yu2020searching}
Z.~Yu, C.~Zhao, Z.~Wang, Y.~Qin, Z.~Su, X.~Li, F.~Zhou, and G.~Zhao, ``Searching central difference convolutional networks for face anti-spoofing,'' in \emph{Proceedings of the IEEE/CVF Conference on Computer Vision and Pattern Recognition}, 2020, pp. 5295--5305.

\bibitem{chen2024dea}
Z.~Chen, Z.~He, and Z.-M. Lu, ``{DEA-Net}: Single image dehazing based on detail-enhanced convolution and content-guided attention,'' \emph{IEEE Transactions on Image Processing}, vol.~33, pp. 1002--1015, 2024.

\bibitem{peng2024aigc}
F.~Peng, H.~Fu, A.~Ming, C.~Wang, H.~Ma, S.~He, Z.~Dou, and S.~Chen, ``{AIGC} image quality assessment via image-prompt correspondence,'' in \emph{Proceedings of the IEEE/CVF Conference on Computer Vision and Pattern Recognition}, 2024, pp. 6432--6441.

\bibitem{ganin2015unsupervised}
Y.~Ganin and V.~Lempitsky, ``Unsupervised domain adaptation by backpropagation,'' in \emph{International Conference on Machine Learning}.\hskip 1em plus 0.5em minus 0.4em\relax PMLR, 2015, pp. 1180--1189.

\end{thebibliography}

\end{document}


\title{Supplementary Material}

\maketitle

\gr
\section{Overview}
\label{sec:overview}

This supplementary material provides empirical and conceptual evidence to support the main manuscript. We structure the extended analyses to address architectural mechanisms, statistical reliability, and generalization capabilities across diverse visual domains. Specifically, Section B extends our evaluation to large-scale benchmarks to confirm the scalability of our approach against modern generative priors. Section C presents dataset-level and network-level statistical validations of the interaction hypothesis. Section D provides visual analyses of the specialized difference convolution experts and their dynamic routing behavior. Section E establishes the mathematical rigor of our improvements through statistical significance analysis. Finally, Section F verifies the feature disentanglement mechanism through cross-task feature decoupling and demonstrates our implicit anomaly localization capabilities.

\section{Extended Evaluation on Contemporary Benchmarks}
\label{sec:extended_evaluation}

\vspace{1ex}
\noindent\textbf{Performance on Fine-Grained Contemporary Datasets.} To further demonstrate scalability, we evaluated ACL-IQA on contemporary datasets focusing predominantly on fine grained alignment and compositional artifacts, including EvalMuse-40K~\cite{han2024evalmuse}, GenAIBench~\cite{li2024genai}, and RichHF-18K~\cite{liang2024rich}. Specifically, EvalMuse-40K provides 40,000 image and text pairs with comprehensive human annotations evaluating alignment. GenAIBench focuses on compositional generation by utilizing highly complex prompts involving logic and attribute binding. RichHF-18K contains 18,000 images annotated with rich human feedback highlighting local artifacts and semantic mismatches. For EvalMuse and GenAIBench, we bypassed the dual-path mechanism in favor of a single-path architecture with adaptive differential convolutions~\cite{su2021pixel, kong2024moe}. For RichHF-18K, we utilized its provided artifact scores to drive our dual-path mechanism. Regarding the experimental setup, we adopted distinct data splitting protocols based on availability. Because the official test labels for EvalMuse-40K are not publicly released, we partitioned the available data into a four to one train and test split while maintaining a consistent score distribution, resulting in 26,173 training samples and 6,544 testing samples. For GenAIBench and RichHF-18K, we directly utilized the official splits provided by the dataset authors. To facilitate unified processing, we linearly mapped all subjective scores to a standard range from zero to five. During optimization, we trained our model for 100 epochs on these three datasets.

\begin{table}[htp]
\centering
\caption{Performance comparison measured by SRCC and PLCC on EvalMuse-40K, GenAIBench, and RichHF-18K focusing on Alignment quality. The FT column indicates whether the model was fine-tuned. Both the best and second-best results are highlighted in bold.}
\label{tab:alignment_only_supp}
\resizebox{0.48\textwidth}{!}{
\begin{tabular}{l|c|c|c|c}
\noalign{\global\arrayrulewidth=1.0pt}\hline\noalign{\global\arrayrulewidth=0.4pt}
\rowcolor{gray!15}
Method & FT & EvalMuse & GenAIBench & RichHF \\
\hline
LIQE~\cite{zhang2023blind} & $\checkmark$ & 0.7708 / 0.7583 & 0.5528 / 0.5513 & 0.6884 / 0.6616 \\
FGA-BLIP2~\cite{han2024evalmuse} & $\checkmark$ & 0.7723 / 0.7716 & 0.5639 / 0.5578 & 0.6548 / 0.6415 \\
Q-Align~\cite{wu2024qalign} & \ding{55} & 0.3236 / 0.3241 & 0.2539 / 0.2275 & 0.2654 / 0.2774 \\
Q-Align~\cite{wu2024qalign} & $\checkmark$ & 0.7467 / 0.7375 & 0.6536 / \textbf{0.6365} & 0.7120 / 0.7071 \\
LMM4LMM~\cite{wang2025lmm4lmm} & \ding{55} & 0.7284 / 0.6638 & \textbf{0.7445} / 0.5870 & 0.7193 / 0.6589 \\
LMM4LMM~\cite{wang2025lmm4lmm} & $\checkmark$ & \textbf{0.8102} / \textbf{0.8038} & \textbf{0.7459} / \textbf{0.7244} & \textbf{0.7485} / \textbf{0.7293} \\
\hline
\rowcolor{green!8!white}
\textbf{ACL-IQA} & $\checkmark$ & \textbf{0.7908} / \textbf{0.7873} & 0.6022 / 0.5901 & \textbf{0.7324} / \textbf{0.7152} \\
\noalign{\global\arrayrulewidth=1.0pt}\hline\noalign{\global\arrayrulewidth=0.4pt}
\end{tabular}
}
\end{table}

As shown in Table~\ref{tab:alignment_only_supp}, ACL-IQA outperforms strong vision-language baselines like LIQE~\cite{zhang2023blind} and FGA-BLIP2~\cite{han2024evalmuse} and supervised fine-tuned MLLMs like Q-Align~\cite{wu2024qalign} across EvalMuse and RichHF-18K, achieving an impressive correlation on EvalMuse and RichHF. While parameter-heavy MLLMs of approximately 8B parameters exhibit an advantage on GenAIBench due to deep semantic logic reasoning on compositional prompts, our lightweight model of approximately 192M parameters delivers highly competitive and computationally efficient evaluations without relying on resource-intensive architectures.

\vspace{1ex}
\noindent\textbf{Generative Prior Domain Shift Analysis.} To illustrate the substantial gap between early-stage datasets, such as AGIQA-3K~\cite{li2023agiqa} and AIGCIQA2023~\cite{wang2023aigciqa2023}, and contemporary benchmarks like EvalMi-50K~\cite{wang2025lmm4lmm}, we performed t-SNE visualizations on textual prompt embeddings extracted via Qwen3-Embedding-0.6B~\cite{zhang2025qwen3} and visual image features extracted via CLIP ViT-B/32~\cite{radford2021learning}.

\begin{figure}[htp]
\centering
\includegraphics[width=0.85\linewidth]{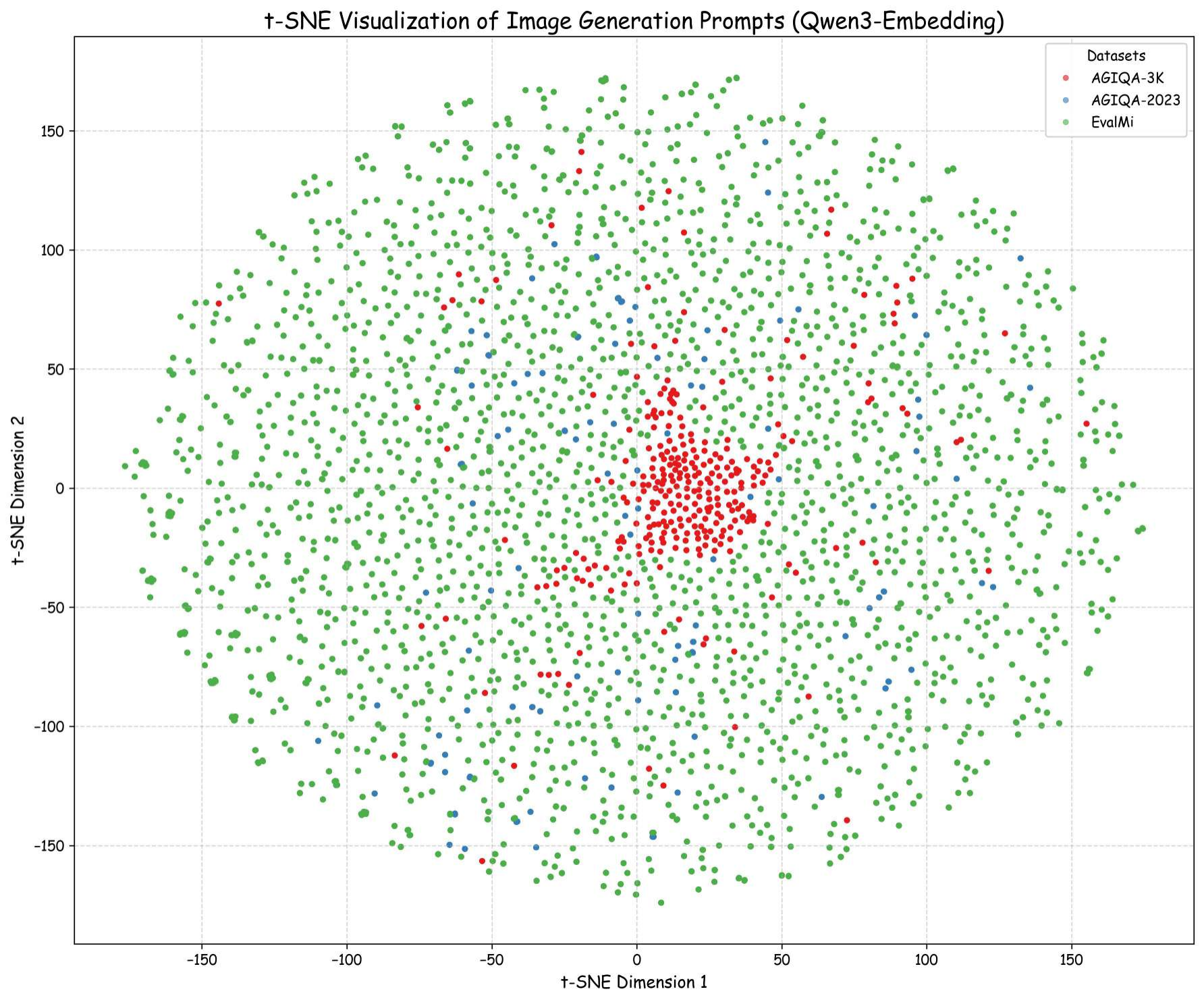}
\caption{t-SNE visualization of textual prompt embeddings extracted via Qwen3-Embedding-0.6B~\cite{zhang2025qwen3}, demonstrating the expanded semantic manifold of contemporary benchmarks.}
\label{fig:tsne_prompts}
\end{figure}

\begin{figure}[htp]
\centering
\includegraphics[width=0.85\linewidth]{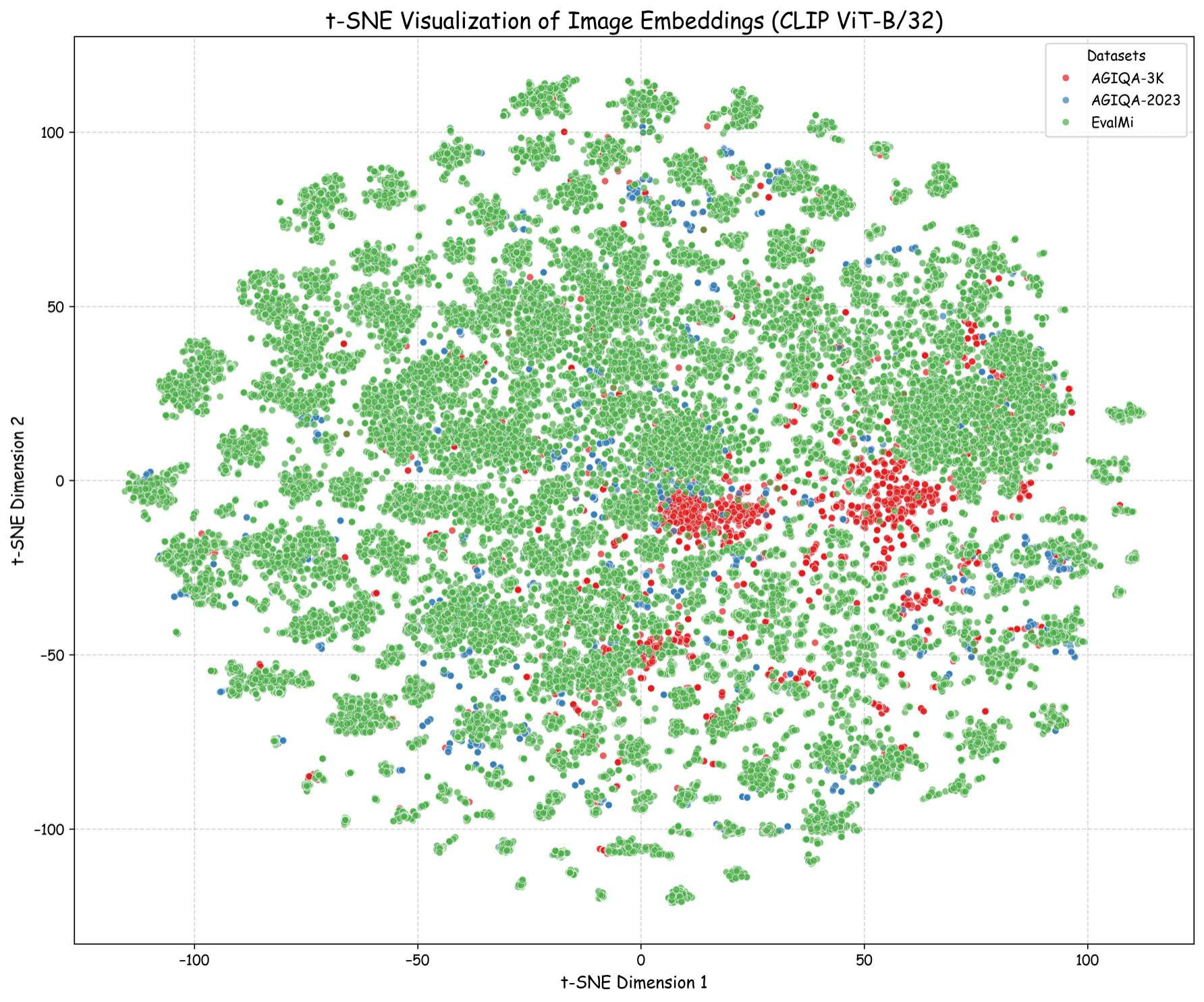}
\caption{t-SNE visualization of generated image features extracted via CLIP ViT-B/32~\cite{radford2021learning}, illustrating the pronounced visual distribution shift in modern generators.}
\label{fig:tsne_images}
\end{figure}

As shown in Figs.~\ref{fig:tsne_prompts} and \ref{fig:tsne_images}, modern generators in EvalMi-50K encompass a dramatically wider and more dispersed semantic and visual manifold compared to the tightly clustered spaces of early-stage datasets. Despite this pronounced distribution shift, our cross-dataset transfer experiments confirm that ACL-IQA successfully bridges these domain discrepancies, proving that our interaction-aware mechanism learns fundamental quality dynamics rather than overfitting to early generative artifacts.

\section{Conceptual Validation of Adversarial and Collaborative Interactions}
\label{sec:conceptual_validation}

To  substantiate that the collaborative and adversarial interactions between perceptual fidelity and prompt alignment represent a universal phenomenon across AI-Generated Image Quality Assessment, also referred to as AIGIQA, rather than incidental edge cases, we provide comprehensive statistical validations across both macroscopic dataset distributions and microscopic network optimization dynamics.

\vspace{1ex}
\noindent\textbf{Prompt-Level Semantic Quadrant Analysis.} Evaluating the correlation between perception and alignment across an entire dataset without grouping yields a misleadingly high positive correlation due to the massive performance gap between early and modern generative models. To eliminate this confounding model-capacity variance and uncover the true semantic distribution, we treat each textual prompt as an independent semantic unit. Specifically, we compute the mean perception quality and mean alignment quality for all images generated under each specific prompt across AGIQA-3K, AIGCIQA2023, RichHF-18K, and EvalMi-50K.

\begin{figure}[htp]
\centering
\includegraphics[width=1\linewidth]{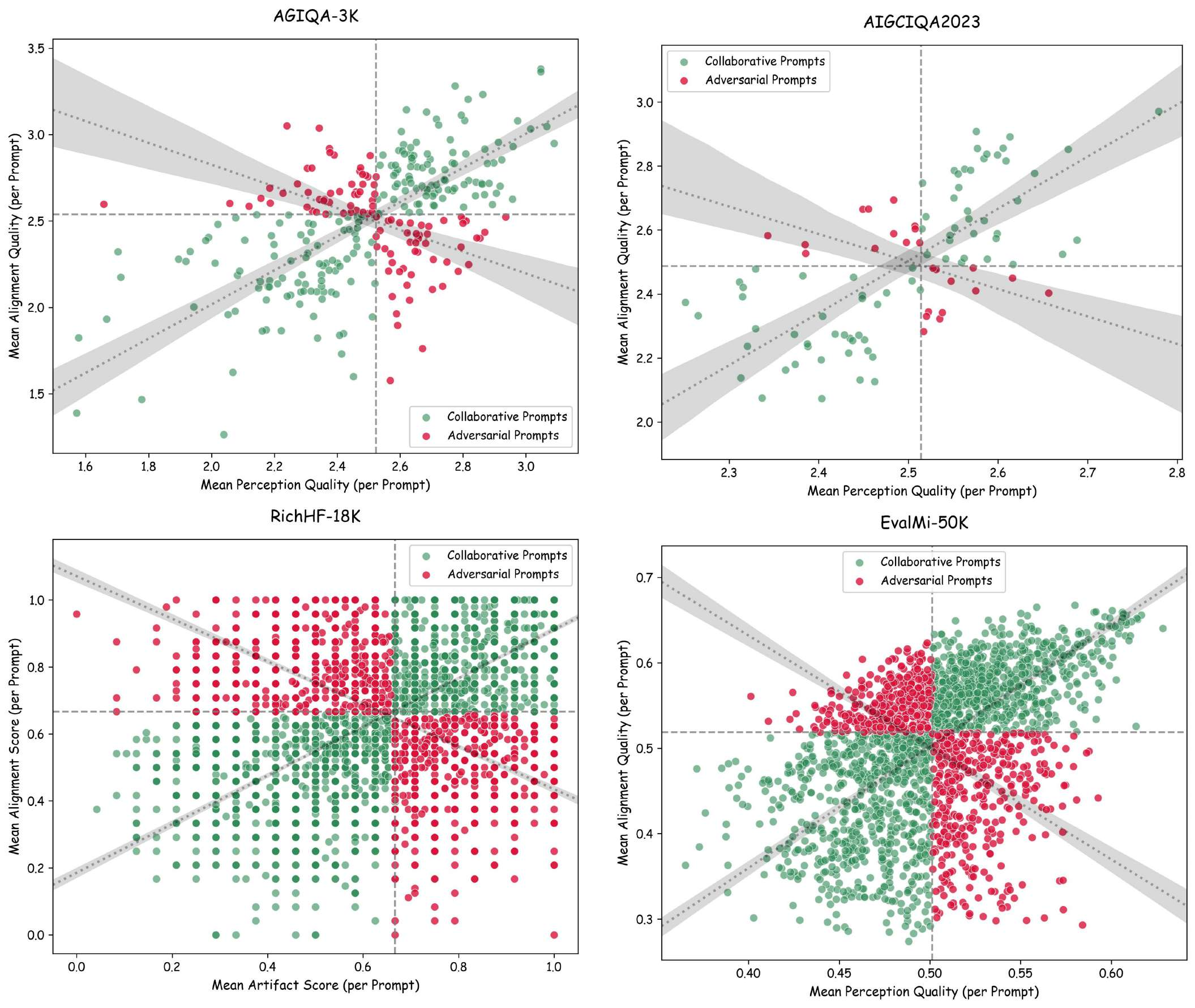}
\caption{Prompt-level Semantic Quadrant Analysis across AGIQA-3K, AIGCIQA2023, RichHF-18K, and EvalMi-50K datasets. Each point represents the mean perception and alignment scores of all images generated under a specific prompt by various models. The distribution explicitly delineates collaborative scenarios in the first and third quadrants and adversarial scenarios in the second and fourth quadrants. The regression line and shaded band represent the global positive data trend and its 95\% confidence interval, while intersecting lines denote dataset medians.}
\label{fig:prompts_quadrant}
\end{figure}

As illustrated in Fig.~\ref{fig:prompts_quadrant}, partitioning the prompts into four quadrants based on dataset medians reveals a distinct bifurcation. The first and third quadrants represent collaborative scenarios, where simple or standard prompts allow models to succeed or fail at both tasks simultaneously, forming a mutually synergistic relationship. Conversely, the second and fourth quadrants represent adversarial scenarios, which include highly complex, geometrically constrained, or surreal instructions. Under these strict semantic constraints, generative models are forced into an inherent trade-off, which means they either aggressively satisfy textual constraints at the expense of visual realism to produce high alignment and low perception, or they prioritize visual aesthetics while ignoring complex prompt details to yield high perception and low alignment. Statistically, these adversarial prompts occupy a highly significant portion of modern datasets, accounting for 41.40\% of the 6,846 total prompts in RichHF-18K, 36.56\% of the 2,090 total prompts in EvalMi-50K, 33.33\% in AGIQA-3K, and 24.00\% in AIGCIQA2023. This confirms that without explicit feature disentanglement, forced alignment representations will severely corrupt perception quality predictions and vice versa.

\vspace{1ex}
\noindent\textbf{Network-Level Gradient Conflict Tracking.} To provide direct proof of this dual interaction during actual model training, we tracked the step-by-step optimization dynamics within the shared parameter space of our architecture. Specifically, we configured the training batch size to 8 to capture fine-grained optimization behaviors without the gradient-smoothing effect inherent in large batches. For each training batch, we extracted the gradient vector of the alignment score loss and the gradient vector of the perception or artifact score loss with respect to the shared backbone features.

\begin{figure}[htp]
\centering
\includegraphics[width=1\linewidth]{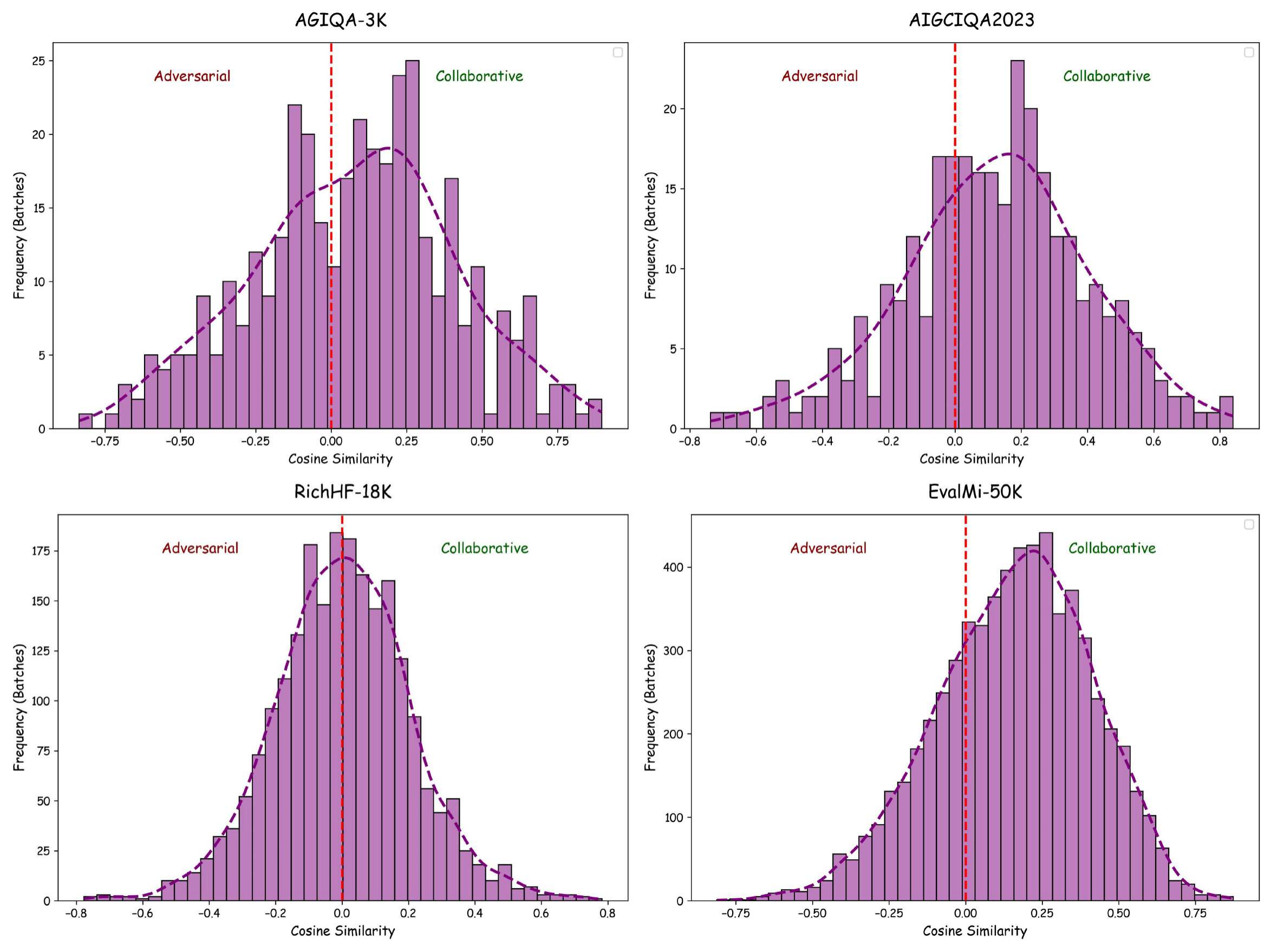}
\caption{Network-level Gradient Conflict Tracking. The histograms display the distribution of batch-wise cosine similarities between the alignment gradients and the perception or artifact gradients during training across the AGIQA-3K, AIGCIQA2023, RichHF-18K, and EvalMi-50K datasets. The curve indicates the Kernel Density Estimation, and the vertical line marks the zero-similarity boundary.}
\label{fig:batch_cosine_similarity}
\end{figure}

As shown by the cosine similarity histograms and Kernel Density Estimation curves in Fig.~\ref{fig:batch_cosine_similarity}, the distribution explicitly bifurcates across the zero-similarity boundary across all benchmarks. A significant mass resides in the positive domain, where the cosine similarity is greater than zero, mathematically proving synergistic collaborative learning where tasks share consistent optimization directions. Crucially, an equally massive portion of the distribution falls into the negative domain, where the cosine similarity is less than zero. This destructive interference provides direct evidence of severe gradient conflicts, demonstrating that alignment optimization directly opposes prediction optimization for complex semantic batches. This empirical evidence firmly validates the necessity of our Gradient Reversal Layer, or GRL, for adversarial decoupling and the Dual-Gated Mixture-of-Experts, known as MoE, dynamic routing.

\section{Visualization of Specialized Convolutional Experts}
\label{sec:expert_visualization}

Our framework embeds difference convolutions into the deep latent space of the Vision Transformer~\cite{dosovitskiy2020image} to capture fine-grained details in AI-generated images. We employ six types of convolution experts: Vanilla, Central Difference~\cite{yu2020searching}, Angular Difference, Radial Difference, Horizontal Difference, and Vertical Difference Convolutions\cite{su2021pixel, chen2024dea}.

\begin{figure*}[htp]
    \centering
    \includegraphics[width=0.8\textwidth]{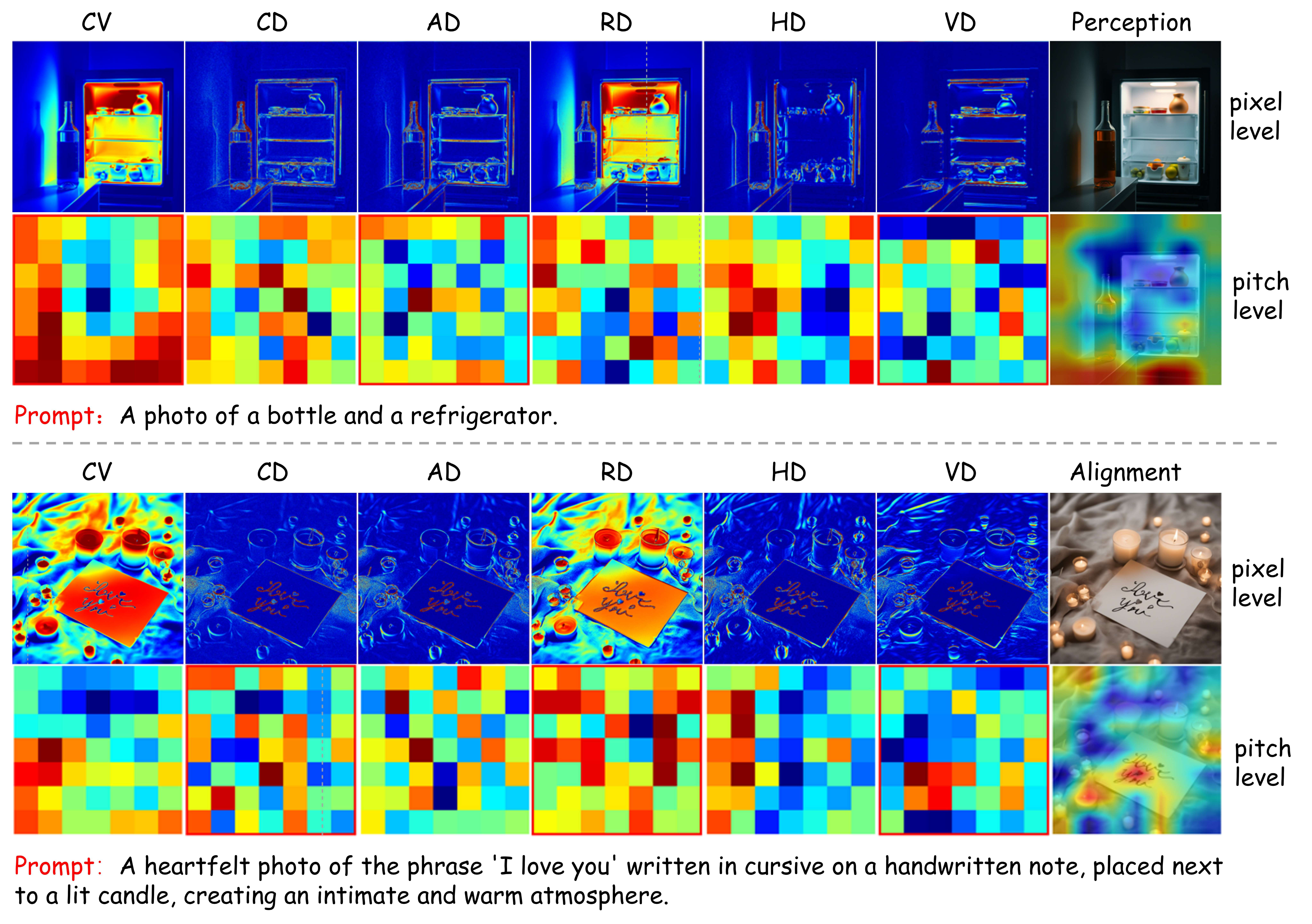}
    \caption{Feature maps from convolution experts for perception and alignment quality prediction. Each section displays pixel-level and patch-level features. Red boxes denote the top-3 dynamically gated experts, which are fused to form the final map, involving Vanilla Convolution, Central Difference Convolution, Angular Difference Convolution, Radial Difference Convolution, Horizontal Difference Convolution, and Vertical Difference Convolution.}
    \label{fig:kernel_feature_map}
\end{figure*}

\vspace{1ex}
 As visualized in Fig.~\ref{fig:kernel_feature_map}, the intermediate feature maps exhibit highly specialized response patterns. The Vanilla convolution primarily anchors global contextual information and salient objects. In contrast, difference convolutions strongly activate around structural boundaries, misaligned edges, and complex geometric distortions. These operators are embedded deep within the Vision Transformer architecture, each token already possesses a global receptive field, and thus the feature responses present a blocky, patch-wise distribution. Consequently, these operators act upon abstract semantic tokens to capture patch-level semantic or structural discontinuities, enabling the detection of higher-order compositional errors and semantic misalignments rather than low-level pixel noise.

\section{Statistical Significance and Sensitivity Analyses}
\label{sec:statistical_analyses}

\vspace{1ex}
\noindent\textbf{Statistical Significance Analysis.} To ensure that our numerical gains stem from intrinsic algorithmic superiority rather than random data variance, we conducted a rigorous 10-fold cross-validation on AGIQA-3K and AIGCIQA2023, comparing ACL-IQA directly against the competitive baseline IPCE~\cite{peng2024aigc} under identical data splits. We computed the 95\% Confidence Intervals for both metrics alongside paired $t$-tests across the 10 independent evaluation folds.

\begin{table*}[htp]
\centering
\caption{Statistical Significance Analysis using 10-Fold Cross-Validation comparing ACL-IQA against the competitive baseline IPCE. The $p$-values are computed using a paired $t$-test, where a value less than 0.01 indicates an extremely significant improvement.}
\label{tab:significance}
\resizebox{1\textwidth}{!}{
\begin{tabular}{l|l|l|c|c|c}
\noalign{\global\arrayrulewidth=1.0pt}\hline\noalign{\global\arrayrulewidth=0.4pt}
\rowcolor{gray!15}
Dataset & Task & Model & SRCC 95\% CI & PLCC 95\% CI & Paired $t$-test ($p$-value) \\
\hline
\multirow{4}{*}{AGIQA-3K} 
& \multirow{2}{*}{Perception} 
& IPCE & $[0.8775, 0.8907]$ & $[0.9213, 0.9279]$ & \multirow{2}{*}{\begin{tabular}[c]{@{}c@{}}\textbf{$p < 0.01$} \\ (SRCC: 0.0031, PLCC: 0.0040)\end{tabular}} \\
& & \textbf{ACL-IQA} & $\mathbf{[0.8840, 0.8966]}$ & $\mathbf{[0.9249, 0.9316]}$ & \\
\cline{2-6}
& \multirow{2}{*}{Alignment} 
& IPCE & $[0.7563, 0.7831]$ & $[0.8624, 0.8826]$ & \multirow{2}{*}{\begin{tabular}[c]{@{}c@{}}\textbf{$p < 0.01$} \\ (SRCC: 0.0069, PLCC: 0.0018)\end{tabular}} \\
& & \textbf{ACL-IQA} & $\mathbf{[0.7694, 0.7969]}$ & $\mathbf{[0.8698, 0.8902]}$ & \\
\hline
\multirow{4}{*}{AIGCIQA2023} 
& \multirow{2}{*}{Perception} 
& IPCE & $[0.8562, 0.8718]$ & $[0.8736, 0.8840]$ & \multirow{2}{*}{\begin{tabular}[c]{@{}c@{}}\textbf{$p < 0.01$} \\ (SRCC: 0.0094, PLCC: 0.0013)\end{tabular}} \\
& & \textbf{ACL-IQA} & $\mathbf{[0.8650, 0.8790]}$ & $\mathbf{[0.8806, 0.8925]}$ & \\
\cline{2-6}
& \multirow{2}{*}{Alignment} 
& IPCE & $[0.7887, 0.8071]$ & $[0.7792, 0.7982]$ & \multirow{2}{*}{\begin{tabular}[c]{@{}c@{}}\textbf{$p < 0.001$} \\ (SRCC: 0.0006, PLCC: 0.0001)\end{tabular}} \\
& & \textbf{ACL-IQA} & $\mathbf{[0.8004, 0.8181]}$ & $\mathbf{[0.7919, 0.8091]}$ & \\
\noalign{\global\arrayrulewidth=1.0pt}\hline\noalign{\global\arrayrulewidth=0.4pt}
\end{tabular}
}
\end{table*}

As demonstrated in Table~\ref{tab:significance}, our proposed ACL-IQA successfully passes the statistical significance test across all evaluations. Not only does our model achieve higher confidence intervals, but the $p$-values generated by the paired $t$-test are all well below 0.01, conclusively verifying that our decoupled interaction-aware mechanism brings a mathematically significant improvement rather than a coincidental random fluctuation.

\vspace{1ex}
\noindent\textbf{Prompt Sensitivity Analysis.} To validate the necessity and robustness of our text template formulation, we systematically designed four distinct variants: Default, Concise, Verbose, and Synonym templates. When conceiving these variations, we deliberately manipulated key linguistic dimensions including sentence structure, instructional verbosity, and synonym phrasing. Specifically, the concise variant stripped away all decorative words to test minimalism, the verbose variant expanded the instructions into complex compound sentences to test long-context attention, and the synonym variant replaced core task keywords with semantically equivalent phrasing to test lexical sensitivity.

We evaluated training sensitivity by retraining the model from scratch on the EvalMi-50K dataset for 10 epochs using each of the four conceptual variations. As summarized in Table~\ref{tab:prompt_sensitivity}, our gating mechanism remains highly robust across different language formulations with generally stable convergence. Crucially, our carefully designed Default Template optimally guides the MoE routing, achieving the highest overall balance and correlation for both perception and alignment tasks. Furthermore, during inference, evaluating unseen templates on a trained model yielded virtually negligible performance fluctuations due to the underlying hard-routing stability of our cross-attention mechanism.

\begin{table}[htp]
\centering
\caption{Performance comparison using SRCC and PLCC of different prompt templates when trained on the EvalMi-50K dataset.}
\label{tab:prompt_sensitivity}
\resizebox{0.48\textwidth}{!}{
\begin{tabular}{l|cc}
\noalign{\global\arrayrulewidth=1.0pt}\hline\noalign{\global\arrayrulewidth=0.4pt}
\rowcolor{gray!15}
Prompt Template & Perception & Alignment \\
\hline
Default Template & \textbf{0.8719} / \textbf{0.8982} & \textbf{0.8556} / \textbf{0.8673} \\
Concise Template & 0.8705 / 0.8985 & 0.8545 / 0.8660 \\
Verbose Template & 0.8719 / 0.8991 & 0.8551 / 0.8672 \\
Synonym Template & 0.8693 / 0.8978 & 0.8531 / 0.8639 \\
\noalign{\global\arrayrulewidth=1.0pt}\hline\noalign{\global\arrayrulewidth=0.4pt}
\end{tabular}
}
\end{table}

\section{Cross-Task Feature Decoupling and Implicit Localization}
\label{sec:feature_decoupling}

\vspace{1ex}
\noindent\textbf{Cross-Task Feature Decoupling Evaluation.} To directly verify the claimed relationships and the exact efficacy of our disentanglement mechanism, we designed a cross-task feature decoupling evaluation across AGIQA-3K, AIGCIQA2023, and EvalMi-50K. By isolating the collaborative learning path $F_{col}$ and the adversarial learning path $F_{adv}$, which are extracted from fully trained models, we forced these decoupled features to independently predict both Perception MOS and Alignment MOS.

\begin{table*}[htp]
\centering
\caption{Cross task feature decoupling evaluation for the ACL-IQA model trained to predict Perceptual Quality. $F_{col}$ denotes the Collaborative Feature, and $F_{adv}$ denotes the Adversarial Feature. The symbol $\checkmark$ indicates the active feature used for prediction, while the downward arrow indicates the relative performance decrease compared to the collaborative path.}
\label{tab:cross_val_perc}
\resizebox{1\textwidth}{!}{
\begin{tabular}{c|cc|cc|cc}
\noalign{\global\arrayrulewidth=1.0pt}\hline\noalign{\global\arrayrulewidth=0.4pt}
\rowcolor{gray!15}
Dataset & $F_{col}$ & $F_{adv}$ & SRCC for Perception & PLCC for Perception & SRCC for Alignment & PLCC for Alignment \\
\hline
\multirow{3}{*}{AGIQA-3K} 
& $\checkmark$ & $\checkmark$ & \textbf{0.8969} & \textbf{0.9162} & - & - \\
\cline{2-7}
& $\checkmark$ & & 0.8819 & 0.8858 & 0.7207 & 0.8125 \\
\cline{2-7}
& & $\checkmark$ & 0.8282 & 0.7777 & 0.3757 \, \textbf{\textcolor{red}{($\downarrow$ 47.9\%)}} & 0.0299 \, \textbf{\textcolor{red}{($\downarrow$ 96.3\%)}} \\
\hline
\multirow{3}{*}{AIGCIQA2023} 
& $\checkmark$ & $\checkmark$ & \textbf{0.8720} & \textbf{0.8865} & - & - \\
\cline{2-7}
& $\checkmark$ & & 0.8379 & 0.8588 & 0.7564 & 0.7508 \\
\cline{2-7}
& & $\checkmark$ & 0.8403 & 0.8576 & -0.0427 \, \textbf{\textcolor{red}{($\downarrow$ 105.6\%)}} & 0.0438 \, \textbf{\textcolor{red}{($\downarrow$ 94.2\%)}} \\
\hline
\multirow{3}{*}{EvalMi-50K} 
& $\checkmark$ & $\checkmark$ & \textbf{0.8810} & \textbf{0.9050} & - & - \\
\cline{2-7}
& $\checkmark$ & & 0.7697 & 0.7955 & 0.7048 & 0.7423 \\
\cline{2-7}
& & $\checkmark$ & 0.7792 & 0.8251 & 0.0379 \, \textbf{\textcolor{red}{($\downarrow$ 94.6\%)}} & -0.0159 \, \textbf{\textcolor{red}{($\downarrow$ 102.1\%)}} \\
\noalign{\global\arrayrulewidth=1.0pt}\hline\noalign{\global\arrayrulewidth=0.4pt}
\end{tabular}
}
\end{table*}

\begin{table*}[htp]
\centering
\caption{Cross task feature decoupling evaluation for the ACL-IQA model trained to predict Alignment Quality. The table presents the joint performance alongside isolated feature predictions, where the downward arrow indicates the relative performance decrease compared to the collaborative path.}
\label{tab:cross_val_align}
\resizebox{1\textwidth}{!}{
\begin{tabular}{c|cc|cc|cc}
\noalign{\global\arrayrulewidth=1.0pt}\hline\noalign{\global\arrayrulewidth=0.4pt}
\rowcolor{gray!15}
Dataset & $F_{col}$ & $F_{adv}$ & SRCC for Alignment & PLCC for Alignment & SRCC for Perception & PLCC for Perception \\
\hline
\multirow{3}{*}{AGIQA-3K} 
& $\checkmark$ & $\checkmark$ & \textbf{0.7832} & \textbf{0.8800} & - & - \\
\cline{2-7}
& $\checkmark$ & & 0.7161 & 0.8116 & 0.8788 & 0.8971 \\
\cline{2-7}
& & $\checkmark$ & 0.7336 & 0.8248 & 0.2194 \, \textbf{\textcolor{red}{($\downarrow$ 75.0\%)}} & 0.2393 \, \textbf{\textcolor{red}{($\downarrow$ 73.3\%)}} \\
\hline
\multirow{3}{*}{AIGCIQA2023} 
& $\checkmark$ & $\checkmark$ & \textbf{0.8092} & \textbf{0.8005} & - & - \\
\cline{2-7}
& $\checkmark$ & & 0.7257 & 0.7213 & 0.8469 & 0.8731 \\
\cline{2-7}
& & $\checkmark$ & 0.7653 & 0.7655 & 0.4242 \, \textbf{\textcolor{red}{($\downarrow$ 49.9\%)}} & 0.3691 \, \textbf{\textcolor{red}{($\downarrow$ 57.7\%)}} \\
\hline
\multirow{3}{*}{EvalMi-50K} 
& $\checkmark$ & $\checkmark$ & \textbf{0.8664} & \textbf{0.8795} & - & - \\
\cline{2-7}
& $\checkmark$ & & 0.6578 & 0.6980 & 0.7633 & 0.8092 \\
\cline{2-7}
& & $\checkmark$ & 0.6126 & 0.6370 & 0.3275 \, \textbf{\textcolor{red}{($\downarrow$ 57.1\%)}} & 0.3107 \, \textbf{\textcolor{red}{($\downarrow$ 61.6\%)}} \\
\noalign{\global\arrayrulewidth=1.0pt}\hline\noalign{\global\arrayrulewidth=0.4pt}
\end{tabular}
}
\end{table*}

As shown in Tables~\ref{tab:cross_val_perc} and \ref{tab:cross_val_align}, the collaborative path $F_{col}$ is designed to capture complementary synergy, retaining mutual information and yielding high accuracy when predicting both primary and secondary dimensions. Conversely, when $F_{adv}$ is evaluated on the secondary, adversarially decoupled task, its correlation converges toward near-zero or slightly negative values. This targeted degradation validates that the Gradient Reversal Layer~\cite{ganin2015unsupervised} successfully strips away interfering representations, ensuring that perceptual features remain unbiased by prompt alignment constraints and vice versa.

To visually substantiate how our dual path mechanism adapts to diverse semantics, we conducted a granular qualitative evaluation. As illustrated in Figs.~\ref{fig:qual_perception} and \ref{fig:qual_alignment}, our collaborative path $F_{col}$ excels in collaborative regimes, where visual clarity and semantic alignment are mutually reinforcing. Conversely, in adversarial regimes involving complex constraints or intentional stylistic degradations, relying solely on $F_{col}$ introduces bias by conflating semantic execution with visual quality. Our adversarial path $F_{adv}$ mitigates this by embedding a Gradient Reversal Layer to actively unlearn interfering representations, mimicking human cognitive inhibition to judge each dimension independently. As demonstrated, our joint framework dynamically shifts routing weights via the Dual Gated Mixture of Experts to bridge these paradigms, effectively replicating human adaptability and ensuring robust evaluation reliability across diverse generative content.

\begin{figure*}[htp]
    \centering
    \includegraphics[width=1\textwidth]{Fig/perception.pdf}
    \caption{Qualitative diagnostic panels for Perceptual Quality Assessment. The top rows illustrate the collaborative dominance regime where standard prompts benefit from shared alignment features. The bottom rows illustrate the adversarial dominance regime where complex prompts trigger semantic interference in the collaborative path, necessitating active feature decoupling via the adversarial path. Our joint ACL-IQA model dynamically balances both paths to match ground truth ratings.}
    \label{fig:qual_perception}
\end{figure*}

\begin{figure*}[htp]
    \centering
    \includegraphics[width=1\textwidth]{Fig/alignment.pdf}
    \caption{Qualitative diagnostic panels for Alignment Quality Assessment. The top rows show scenarios where visual clarity is essential to verify fine grained semantic details, allowing the collaborative path to excel. The bottom rows demonstrate cases involving intentional stylistic degradation, where the adversarial path effectively strips away aesthetic bias to accurately evaluate text image correspondence.}
    \label{fig:qual_alignment}
\end{figure*}

\vspace{1ex}
\noindent\textbf{Implicit Anomaly Localization via Grad-CAM.} While our framework is designed as a global scoring architecture without computationally expensive dense segmentation heads, it internally performs precise spatial anomaly localization. To demonstrate this, we conducted a Gradient-weighted Class Activation Mapping analysis on the RichHF-18K dataset by isolating the logits corresponding to the lowest quality levels, representing bad and poor quality, and backpropagating gradients to the deep ViT feature maps.

\begin{figure}[htp]
    \centering
    \includegraphics[width=0.48\textwidth]{Fig/artifact_map.pdf}
    \caption{Visualizations of implicit localization for artifacts and implausibility on RichHF-18K. The columns display the generated image, human-annotated Ground Truth mask, anomaly-targeted gradient heatmap from ACL-IQA, and textual prompt.}
    \label{fig:artifact_map}
\end{figure}

\begin{figure}[htp]
    \centering
    \includegraphics[width=0.48\textwidth]{Fig/alignment_map.pdf}
    \caption{Visualizations of implicit localization for text-image misalignment on RichHF-18K. The columns display the generated image, human-annotated Ground Truth mask, anomaly-targeted gradient heatmap from ACL-IQA, and textual prompt.}
    \label{fig:alignment_map}
\end{figure}

As illustrated in Figs.~\ref{fig:artifact_map} and \ref{fig:alignment_map}, our model successfully locates specific generative flaws without ever being trained on pixel-level mask supervision. For structural distortions in Fig.~\ref{fig:artifact_map}, high-activation heatmaps precisely cover distorted anatomy, such as deformed hands, and structural anomalies, overlapping with human-annotated Ground Truth masks. For text-image misalignment in Fig.~\ref{fig:alignment_map}, the anomaly-targeted gradients accurately pinpoint misspelled rendered text, for instance Panera Bread logos, and missing objects dictated by the prompt. This confirms that our global quality scores are grounded in fine-grained spatial and semantic understanding.

\bibliographystyle{IEEEtran}
\bibliography{supplement}